\documentclass[journal]{IEEEtran}
\usepackage{amsmath}
\usepackage{algorithm}
\usepackage{algorithmic}
\usepackage{graphicx}
\usepackage{multirow}
\usepackage{tabularx}
\usepackage{xcolor}
\usepackage{array}
\usepackage{amssymb}
\usepackage{bbding}
\usepackage{subfigure}
\usepackage{diagbox}
\usepackage{rotating}
\usepackage{booktabs}
\usepackage{cite}
\usepackage{CJK}
\usepackage{overpic}
\usepackage[colorlinks,
            linkcolor=red,
            anchorcolor=blue,
            citecolor=green
            ]{hyperref}

\newcommand{\addFig}[1]{}
\newcommand{\addFigs}[1]{}
\newcommand{\tabincell}[2]{\begin{tabular}{@{}#1@{}}#2\end{tabular}}
\usepackage{subfigure}
\newcommand{\etal}{\textit{et~al}.~}
\newcommand{\ie}{\textit{i}.\textit{e}.~}
\newcommand{\eg}{\textit{e}.\textit{g}.~}

\usepackage{balance}
\usepackage{cleveref}
\crefformat{section}{\S#2#1#3} 
\crefformat{subsection}{\S#2#1#3}
\crefformat{subsubsection}{\S#2#1#3}

\ifCLASSINFOpdf
\else
\fi

\begin{document}
%
\title{Personal Fixations-Based Object Segmentation with\\
Object Localization and Boundary Preservation}
%
%
%

\author{Gongyang~Li,
        Zhi~Liu,~\IEEEmembership{Senior Member,~IEEE},
        Ran~Shi,
        Zheng~Hu,\\
        Weijie~Wei,
        Yong~Wu,
        Mengke~Huang,
        and~Haibin~Ling

\thanks{Gongyang Li, Zhi Liu, Zheng Hu, Weijie Wei, Yong Wu, and Mengke Huang are with Shanghai Institute for Advanced Communication and Data Science, Shanghai University, Shanghai 200444, China, and School of Communication and Information Engineering, Shanghai University, Shanghai 200444, China (email: ligongyang@shu.edu.cn; liuzhisjtu@163.com; huzhen1995@shu.edu.cn; codename1995@shu.edu.cn; yong\_wu@shu.edu.cn; huangmengke@shu.edu.cn).}
\thanks{Ran Shi is with School of Computer Science and Engineering, Nanjing University of Science and Technology, Nanjing 210094, China (email: rshi@njust.edu.cn).}
\thanks{Haibin Ling is with the Department of Computer Science, Stony Brook University, Stony Brook, NY 11794, USA (email: hling@cs.stonybrook.edu).}
\thanks{Corresponding author: Zhi Liu}
}

\markboth{IEEE TRANSACTIONS ON IMAGE PROCESSING}%
{Shell \MakeLowercase{\textit{et al.}}: Bare Demo of IEEEtran.cls for IEEE Journals}

\maketitle

\begin{abstract}
As a natural way for human-computer interaction, fixation provides a promising solution for interactive image segmentation. In this paper, we focus on Personal Fixations-based Object Segmentation (PFOS) to address issues in previous studies, such as the lack of appropriate dataset and the ambiguity in fixations-based interaction. In particular, we first construct a new PFOS dataset by carefully collecting pixel-level binary annotation data over an existing fixation prediction dataset, such dataset is expected to greatly facilitate the study along the line.
Then, considering characteristics of personal fixations, we propose a novel network based on Object Localization and Boundary Preservation (OLBP) to segment the gazed objects.
Specifically, the OLBP network utilizes an Object Localization Module (OLM) to analyze personal fixations and locates the gazed objects based on the interpretation.
Then, a Boundary Preservation Module (BPM) is designed to introduce additional boundary information to guard the completeness of the gazed objects.
Moreover, OLBP is organized in the mixed bottom-up and top-down manner with multiple types of deep supervision.
Extensive experiments on the constructed PFOS dataset show the superiority of
the proposed OLBP network over 17 state-of-the-art methods, and demonstrate the effectiveness of the proposed OLM
and BPM components.
The constructed PFOS dataset and the proposed OLBP network are available at
\url{https://github.com/MathLee/OLBPNet4PFOS}.
\end{abstract}


\IEEEpeerreviewmaketitle

\section{Introduction}
\label{sec:Intro}
\IEEEPARstart{F}{ixation} is a flexible interaction mechanism of the human visual system.
Compared with scribble, click and bounding box, fixation provides the most convenient interaction for patients with hand disability, amyotrophic lateral sclerosis (ALS) and polio.
This kind of eye control interaction, \ie fixation, can greatly improve the interaction efficiency
of these patients.
In addition, fixation is closely related to personal information such as
age~\cite{Olivier2017Age,TCDS2018Age} and gender~\cite{Gender2008,Gender2010}.
This means that different individuals may have different perceptions and preferences of a
scene~\cite{Access2018PersonalSal,Xu2019PersonalSal}.
Thus motivated, in this paper, we pay close attention to personal fixations-based object segmentation,
which is a more natural manner for interactive image segmentation.
%

\begin{figure}
\centering
\footnotesize
  \begin{overpic}[width=1\columnwidth]{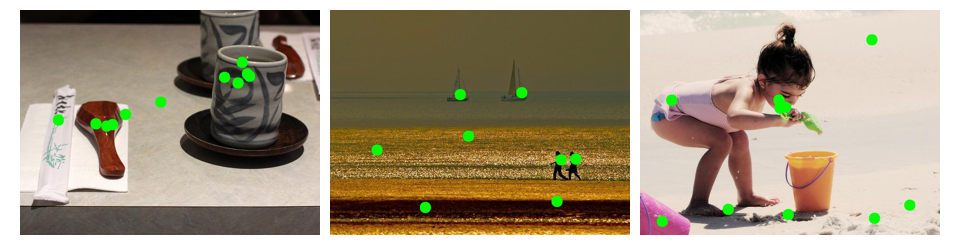}
  \end{overpic}
\caption{Examples of image with ambiguous fixations.
\textcolor{green}{Green dots} in each image indicate fixations.
Some fixations fall in the background.
}
\label{Fig1_background}
\end{figure}

The typical manners of interaction, such as
scribbles~\cite{GraphCut2001,RandomWalk2006,GM2009,CAC2012,HA2014},
clicks~\cite{DIOS2016,RII2017,LZW2018ISLD,18ITIS,BRS2019,TSFN2019}
and bounding boxes~\cite{GrabCut2004,BBP2009,DeepGC2017,ShiRan2018}
for interactive image segmentation,
are explicit behaviors without interference.
By contrast, 
fixations are implicit~\cite{TPAMI2012AVS,2014SOS,SR2017GBOS,LGY2019CFPS}, and their convenience comes with interaction ambiguity.
Concretely, the positive and negative labels of scribbles and clicks are deterministic.
However, fixations are unlabeled when collected.
They do not distinguish between positive labels and negative labels (\ie some fixations may fall in the background as shown in Fig.~\ref{Fig1_background}), resulting in a few noise in the fixations.
Such ambiguous interaction makes the fixations-based object segmentation task difficult. 
Recently, with the rise of convolutional neural networks (CNNs), the clicks-based
interactive image segmentation has been greatly developed.
Even though fixation points and clicking points are similar to some extent, clicks-based methods~\cite{DIOS2016,RII2017,LZW2018ISLD,BRS2019,TSFN2019}
cannot be directly applied to fixations-based object segmentation.

The above observations suggest that there are two main reasons that limit the development of fixations-based object segmentation.
First, there is not a suitable dataset for the fixations-based object segmentation task, let alone dataset based on the personal fixations.
Second, as aforementioned, the ambiguous representation of fixations makes this
type of interaction difficult to handle by other methods which are based on clicks and scribbles.

To address the first crucial issue, we construct a \textit{Personal Fixations-based Object Segmentation} (PFOS) dataset, which is extended from the fixation prediction dataset OSIE~\cite{OSIE2014}.
The PFOS dataset contains 700 images, and each image has 15 personal fixation maps
collected from 15 subjects with corresponding pixel-level annotations of objects.
To overcome the ambiguity of fixations, we propose an effective network based on
\textit{Object Localization and Boundary Preservation} (OLBP).
The key idea of OLBP is to locate the gazed objects based on the analysis
of fixations, and then the boundary information is introduced to guard the completeness of the gazed objects and to filter the background.

In particular, the overall structure of OLBP network is a mixture of bottom-up and top-down architectures.
To narrow the gap between fixations and objects, we propose the
\textit{Object Localization Module} (OLM) to analyze personal fixations in detail and grasp location information of the gazed objects of different individuals.
Based on the interpretation of location information, OLM modulates CNN features of
image in a bottom-up way.
Moreover, considering that the object location information may involve confusing noise, we propose a
\textit{Boundary Preservation Module} (BPM) to exploit boundary information to enforce
object completeness and filter the background of erroneous localization. BPM is integrated into the top-down prediction.
Both OLMs and BPMs employ deep supervision to further improve the capabilities of
feature representation. 
In this way, the scheme of object localization and boundary preservation is successfully applied to the bottom-up and top-down structure, and the proposed OLBP network greatly promotes the performance of the personal fixations-based object segmentation task.
Experimental results on the challenging PFOS dataset demonstrate that OLBP outperforms 17 state-of-the-art methods under various evaluation metrics.

The contributions of this work are summarized as follows:
\begin{itemize}
\item We construct a new dataset for \textit{Personal Fixations-based Object Segmentation} (PFOS),
which focuses on the natural interaction (\ie fixation).
This dataset contains free-view personal fixations without any constraints, expanding its applicability.
We believe that the PFOS dataset will boost the research of fixations-based human-computer interaction.

\item We propose a novel \textit{Object Localization and Boundary Preservation} (OLBP) network to segment the gazed objects based on personal fixations.
The OLBP network, equipped with the \textit{Object Localization Module} and the \textit{Boundary Preservation Module},
effectively overcomes the difficulties from ambiguous fixations.

\item We conduct extensive experiments to evaluate our OLBP network and other state-of-the-art
methods on the PFOS dataset.
Comprehensive results demonstrate the superiority of our OLBP network,
and also reveal the difficulties and challenges of the constructed PFOS dataset.
\end{itemize}

The rest of the paper is organized as follows:
Sec.~\ref{sec:related} reviews related previous works.
Then, we formulate the PFOS task in Sec.~\ref{Problem Statement}.
After that, in Sec.~\ref{Dataset Construction}, we construct the PFOS dataset.
Sec.~\ref{Proposed Method} presents the proposed OLBP network in detail.
In Sec.~\ref{sec:exp}, we evaluate the performance of the proposed OLBP network
and other methods on the constructed PFOS dataset.
Finally, the conclusion is drawn in Sec.~\ref{sec:con}.


\section{Related Work}
\label{sec:related}
In this section, we first give an overview of previous works of interactive image segmentation
in Sec.~\ref{sec:Interactive_Seg}.
Then, we introduce related works of fixations-based object segmentation in Sec.~\ref{sec:FBOSeg}.
Finally, we review some related works on boundary-aware segmentation in Sec.~\ref{sec:Boundary}.

\subsection{Interactive Image Segmentation}
\label{sec:Interactive_Seg}
\textit{1) Scribbles-based interactive image segmentation.}
Scribble is a traditional manner of interaction.
Most of scribbles-based methods are built on graph structures.
GraphCut~\cite{GraphCut2001} is one of the most representative methods.
It uses the max-flow/min-cut theorem to minimize energy function with
hard constraints (\ie labeled scribbles) and soft constraints.
Grady \etal\cite{RandomWalk2006} adopted the random walk algorithm to assign a label
to each unlabeled pixel based on the predefined seed pixels in discrete space.
In~\cite{GM2009}, Bai \etal proposed a weighted geodesic distance based framework,
which is fast for image and video segmentation and matting.
Nguyen \etal\cite{CAC2012} proposed a convex active contour model to segment objects,
and their results were with smooth and accurate boundary contour.
Spina \etal\cite{HA2014} presented a live markers methodology to reduce the user intervention
for effective segmentation of target objects.
Following the seed propagation strategy, Jian \etal\cite{ACP2016} employed the adaptive
constraint propagation to adaptively propagate the scribbles information into the whole image.
Recently, Wang \etal\cite{PD2019} changed their view on interactive image segmentation and 
formulated it as a probabilistic estimation problem, proposing a pairwise likelihood learning
based framework.
These methods are friendly to clearly defined scribbles, but they cannot solve the ambiguity
of fixations and their inference speed is usually slow.

\textit{2) Clicks-based interactive image segmentation.}
Click is a classical manner of interaction.
It has been deeply studied in the deep learning era.
The positive and negative clicks are transformed into two separate Euclidean distance maps
for network input.
Xu \etal\cite{DIOS2016} directly sent RGB image and two distance maps into a
fully convolutional network.
Liew \etal\cite{RII2017} proposed a two-branch fusion network with global prediction and
local regional refinement.
In addition to the RGB image and distance maps, Li \etal\cite{LZW2018ISLD} included clicks
in their network input and proposed an end-to-end segmentation-selection network.
In~\cite{BRS2019}, Jang \etal introduced the backpropagating refinement scheme to
correct mislabeled locations in the initial segmentation map.
Different from the direct concatenation of RGB image and interaction maps of the above
methods, Hu \etal\cite{TSFN2019} separately input RGB image and interaction maps into
two networks, and designed a fusion network for feature interactions.
CNNs have greatly improved the performance of clicks-based interactive image segmentation,
but when these methods are applied to fixations-based object segmentation, some background regions will be mistakenly segmented.
To address the problem of erroneous localization, we explore the boundary information in our
BPM to filter redundant background regions and guard the gazed object.

\textit{3) Bounding boxes-based interactive image segmentation.}
In a bounding box, the target object and background coexist,
which is different from scribble and click.
Rother \etal\cite{GrabCut2004} extended the graph-cut approach, and segmented object
with a rectangle, namely GrabCut.
To overcome the looseness of the bounding box, Lempitsky \etal\cite{BBP2009} incorporated
the tightness prior into the global energy minimization function as hard constraints
to further completed target object. 
Shi \etal\cite{ShiRan2018} proposed a coarse-to-fine method with region-level and
pixel-level segmentation.
Similar to~\cite{DIOS2016}, Xu \etal\cite{DeepGC2017} transformed the bounding box to
a distance map and concatenated it with the RGB image to input into
an encoder-decoder network.
Although bounding box and fixation are similar
(\ie target object and background coexist in both interactions),
the bounding box-based methods are difficult to transfer to fixations-based object segmentation.

\subsection{Fixations-Based Object Segmentation}
\label{sec:FBOSeg}
Fixation plays an integral role in the human visual system and it is convenient for interaction.
In an early study, Sadeghi~\etal\cite{SPIE2009} constructed an eyegaze-based interactive segmentation system which adopts random walker to segment objects.
Meanwhile, Mishra \etal\cite{TPAMI2012AVS} gave the definition of fixations-based object
segmentation, that is, segmenting regions containing fixation points.
They transformed the image to polar coordinate system, and found the optimal contour to fit
the target object.
Based on the interpretation of visual receptive field, Kootstra \etal\cite{SelectFP2010}
used symmetry to select fixations closer to the center of the object to obtain more complete
segmentation.
Differently, Li \etal\cite{2014SOS} focused on selecting the most salient objects,
and they ranked object proposals based on fixations.
Similar to~\cite{2014SOS}, Shi \etal\cite{SR2017GBOS} analyzed the fixation distribution
and proposed three metrics to evaluate the score of each candidate region.
In~\cite{PointCut2015}, Tian \etal first determined the uninterested regions, and then  used superpixel-based random walk model to segment the gazed objects.
Khosravan \etal\cite{Gaze2Segment2017} integrated fixations into the medical image
segmentation and proposed a Gaze2Segment system.
Li \etal\cite{LGY2019CFPS} constructed a dataset where all fixations fall in objects
(\ie constrained fixations), and proposed a CNN-based model to simulate the human visual
system to segment objects based on fixations.

These studies have promoted the development of fixations-based object
segmentation. However, all the fixations in~\cite{TPAMI2012AVS,SelectFP2010,PointCut2015,LGY2019CFPS}
fall in objects, which are hardly guaranteed in practice. 
These methods~\cite{TPAMI2012AVS,SelectFP2010,PointCut2015,LGY2019CFPS}
will get stuck in the ambiguity of unconstrained fixations, especially
of personal fixations.
For~\cite{2014SOS,SR2017GBOS}, they are based on region proposal and
cannot obtain accurate results.
In summary, the above methods cannot solve the problem of ambiguous fixations, as
shown in Fig.~\ref{Fig1_background}.
In this paper, we take advantage of CNNs, and propose a bottom-up and top-down network
to locate objects and preserve objects' boundaries.
Moreover, we construct a dataset to promote this special direction of interactive image segmentation,
\ie personal fixations-based object segmentation.

\subsection{Boundary-Aware Segmentation}
\label{sec:Boundary}
The boundary/edge-aware segmentation idea is widely-used in salient object detection~\cite{Wang2019AFNet,TIP19FB,ICCV2019SCRN,ICCV2019EGNet} and semantic segmentation~\cite{ICCV2019SS}.
In~\cite{Wang2019AFNet}, Wang~\etal modeled the boundary information as an edge-preserving constraint, and included it as an additional supervision in loss function.
In~\cite{TIP19FB}, Wang~\etal proposed a two-branch network, including boundary and mask sub-networks, for jointly predicting masks of salient objects and detecting object boundaries.
In~\cite{ICCV2019SCRN}, Wu~\etal explored the logical interrelations between binary segmentation and edge maps in a multi-task network, and proposed a cross refinement unit in which the segmentation features and edge features are fused in a cross-task manner.
In~\cite{ICCV2019EGNet}, Zhao~\etal focused on the complementarity between salient edge information and salient object information.
They integrated the local edge information of shallow layers and global location information of deep layers to obtain the salient edge features, and then the edge features were fed to the one-to-one guidance module to fuse the complementary region and edge information.
In~\cite{ICCV2019SS}, Ding~\etal first introduced the boundary information as an additional semantic class to enable the network to be aware of the boundary layout, and then proposed a boundary-aware feature propagation network to control the feature propagation based on the learned boundary information.

In our method, we use the boundary information in two aspects: the multi-task structure (\ie segmentation and boundary predictions) and the \textit{Boundary Preservation Module}.
Different from~\cite{TIP19FB,ICCV2019SCRN}, we integrate the learned boundary map into the prediction network in BPMs to preserve the completeness of the gazed objects, rather than fuse the segmentation features and boundary features.
Compared with~\cite{ICCV2019EGNet}, our segmentation prediction is accompanied by the boundary prediction in a uniform prediction network, and the boundary supervision is employed at multiple scales.
Different from~\cite{ICCV2019SS}, which uses the boundary map to control the region of feature propagation, our method uses the boundary map to filter the background of erroneous localization in features.
In short, our use of boundary information is diverse and in-depth, which is suitable for the personal fixations-based object segmentation task.

\section{Personal Fixations-Based Object Segmentation}
\label{Problem Statement}
\textbf{Problem Statement.}
Given an image ${\bf I}$ and a fixation map ${\bf FM}$ of a person,
personal fixations-based object segmentation aims to segment the gazed objects of
this person according to his/her personal ${\bf FM}$, producing a binary segmentation map.
In general, different individuals generate different fixation maps when observing the
same image, which means that individuals may be interested in different objects.
In other words, segmentation results of different individuals on the same image vary with the observer.
%
So, the special characteristic of this task is that an image has multiple binary segmentation maps due to multiple
fixation maps.
Although the ambiguity of fixations makes this task difficulty, the personal fixation map is the only information that can determine the gazed objects.

\textbf{Applications.}
This task has several meaningful applications.
First, such a convenient manner of interaction is conducive to the development of
special eye-control devices for patients with hand disability, ALS and polio,
facilitating their lives and improving their quality of life.
Second, fixation is advantageous to diagnose certain mental illnesses, such as autism spectrum
disorder (ASD)~\cite{WWJ2019ASD,TMM2019SSD} and
schizophrenia spectrum disorders (SSD)~\cite{EAPCN2019SSD,AICT2017SSD}.
This task understands personal fixations at the object level,
which is helpful to improve the accuracy of disease diagnosis.
For example, patients with ASD prefer to pay attention to background rather than foreground,
so the proportion of foreground in their segmentation results will be less than that of healthy people.

\begin{table}[!t]
\centering
\small
\caption{Categories of fixation map (FM) in the PFOS dataset.
Constrained FM means that all fixations fall in the objects/foreground.
Unconstrained FM represents that some fixations fall in the background.
}
\label{table:PFOS}
  \renewcommand{\arraystretch}{1.4}
  \renewcommand{\tabcolsep}{3.5mm}
\begin{tabular}{c|cc}
\midrule[1pt]  
       PFOS dataset  & Constrained FM & Unconstrained FM \\
\hline
\hline
10,500 & 3,683 (35.1\%)  & 6,817 (64.9\%)  \\
\toprule[1pt]
\end{tabular}
\end{table}

\section{Dataset Construction and Transformation}
\label{Dataset Construction}
Currently, there are many prevalently used datasets for fixation prediction, such as
MIT1003~\cite{2009MIT1003}, OSIE~\cite{OSIE2014} and SALICON~\cite{2015SALICON},
and for interactive image segmentation, such as
GrabCut~\cite{GrabCut2004}, Berkeley~\cite{2010Berkeley} and
PASCAL VOC~\cite{2010PASCALVOC}.
However, there is no dataset for the personal fixations-based object segmentation task.
Considering that it is time-consuming for dataset annotations, we propose
a convenient way to collect suitable data from existing datasets for this task.

Obviously, the PFOS dataset must contain fixation data and pixel-level annotations for objects. Among the existing datasets, some datasets, such as DUTS-OMRON~\cite{2013OMRON}, PASCAL-S~\cite{2014SOS} and OSIE~\cite{OSIE2014}, are potential candidates.
The pixel-level annotations of DUTS-OMRON and PASCAL-S are for salient object detection~\cite{2015SODBenchmark,2019sodsurvey,Liu2014ST}, that is, these annotations only focus on the most visually attractive objects but ignore other objects, which could be fixated by different individuals, in a scene.
Therefore, they are not perfect for constructing a PFOS dataset.
Fortunately, the pixel-level annotations of OSIE have semantic attributes. This means that we can select objects, which the user is interested in, based on personal fixations.
In other words, we can create the pixel-level binary ground truths (GTs) for personal fixations-based
object segmentation.
So, we transform the fixation prediction dataset OSIE to our PFOS dataset.

For each image in the OSIE dataset, it has corresponding fixation maps and semantic GTs
of different subjects.
The detailed steps for dataset transformation are as follows:

1) \textit{Semantic labels collection.}
We get the position of each fixation point from the fixation map, and we collect the semantic label of each position in the corresponding semantic GT.

2) \textit{Semantic labels distillation.}
As mentioned in Sec.~\ref{sec:Intro} and shown in Fig.~\ref{Fig1_background}, some fixation
points fall in the background or the same object.
For semantic labels collected from Step 1, we discard the semantic label ``0" which indicates background.
Then, if there are several same semantic labels, we keep only one.

3) \textit{Binary GT creation.}
Based on the distilled semantic labels from Step 2, we can determine the gazed objects and create the binary GT.
We reserve the regions with the distilled semantic labels in the semantic GT, and set them as foreground.
We set the regions with the other unrelated semantic labels as background.

In this convenient way, we efficiently create the binary GTs and successfully construct the PFOS dataset.
The PFOS dataset retains all 700 images and 10,500 free-view personal fixation maps from the OSIE dataset. In the PFOS dataset, the image resolution is $800\times600$. Each image has 15 personal fixation maps from 15 subjects and the transformed binary GTs.
In the constructed PFOS dataset, there are two categories of fixation maps.
The first category is that all fixations fall in the objects/foreground, \ie the constrained fixation map in~\cite{LGY2019CFPS}.
The second category is that some fixations fall in the background, namely the unconstrained fixation map.
We present the details of them in Tab.~\ref{table:PFOS}.
In our PFOS dataset, the unconstrained fixation maps account for 64.9\% and the constrained fixation maps hold 35.1\%.
The large proportion of unconstrained fixation maps increase the ambiguity of our PFOS dataset and make this dataset challenging.

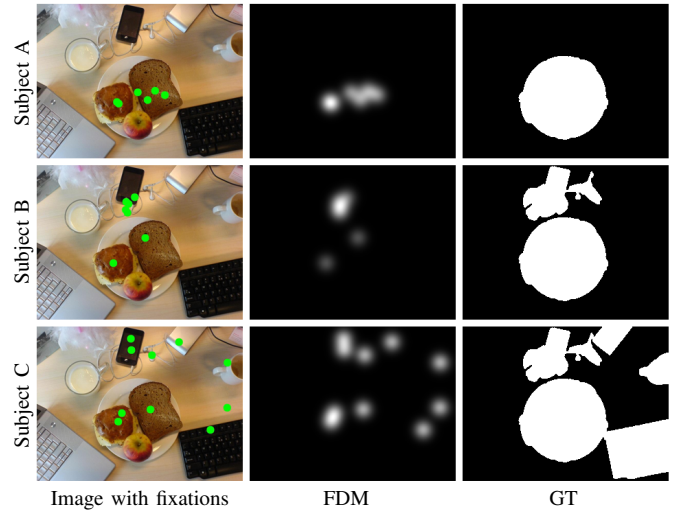
\begin{figure}
\centering
\footnotesize
  \begin{overpic}[width=.998\columnwidth]{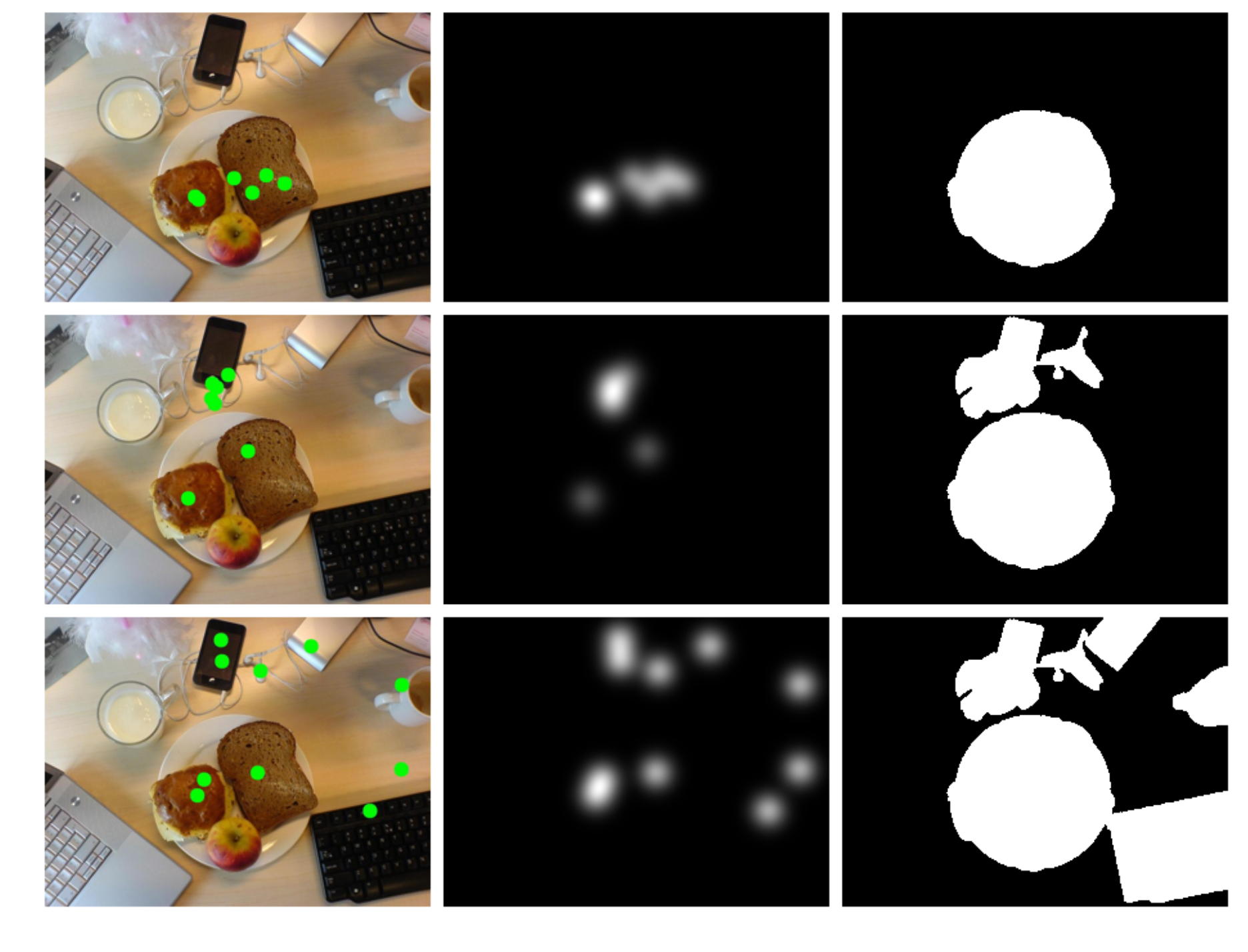}
    \put(4.5,0){   Image with fixations  }
    \put(45.5,0){ FDM }
    \put(79.5,0){ GT }
    \put(-1,57){ \begin{sideways}{Subject A}\end{sideways} }
    \put(-1,33.2){ \begin{sideways}{Subject B}\end{sideways} }
    \put(-1,8.5){ \begin{sideways}{Subject C}\end{sideways} }
  \end{overpic}
\caption{Examples of the PFOS dataset.
\textcolor{green}{Green dots} in each image indicate fixations, FDM is fixation density map, and GT represents ground truth.
}
\label{Fig2_Examples}
\end{figure}

\section{Methodology}
\label{Proposed Method}
In this section, we first conduct data preprocessing which transforms the fixation points into
fixation density maps in Sec.~\ref{sec:FDM}.
Then, we present the overview and motivation of the proposed
\textit{Object Localization and Boundary Preservation} (OLBP) network in Sec.~\ref{Network Overview}.
Next, we give the detailed formulas of the \textit{Object Localization Module} (OLM)
and the \textit{Boundary Preservation Module} (BPM) in Sec.~\ref{Object Localization Module}
and Sec.~\ref{Boundary Preservation Module}, respectively.
Finally, we clarify the implementation details of OLBP network in Sec.~\ref{Implementation Details}.

\begin{figure*}
\centering
\includegraphics[scale=0.65]{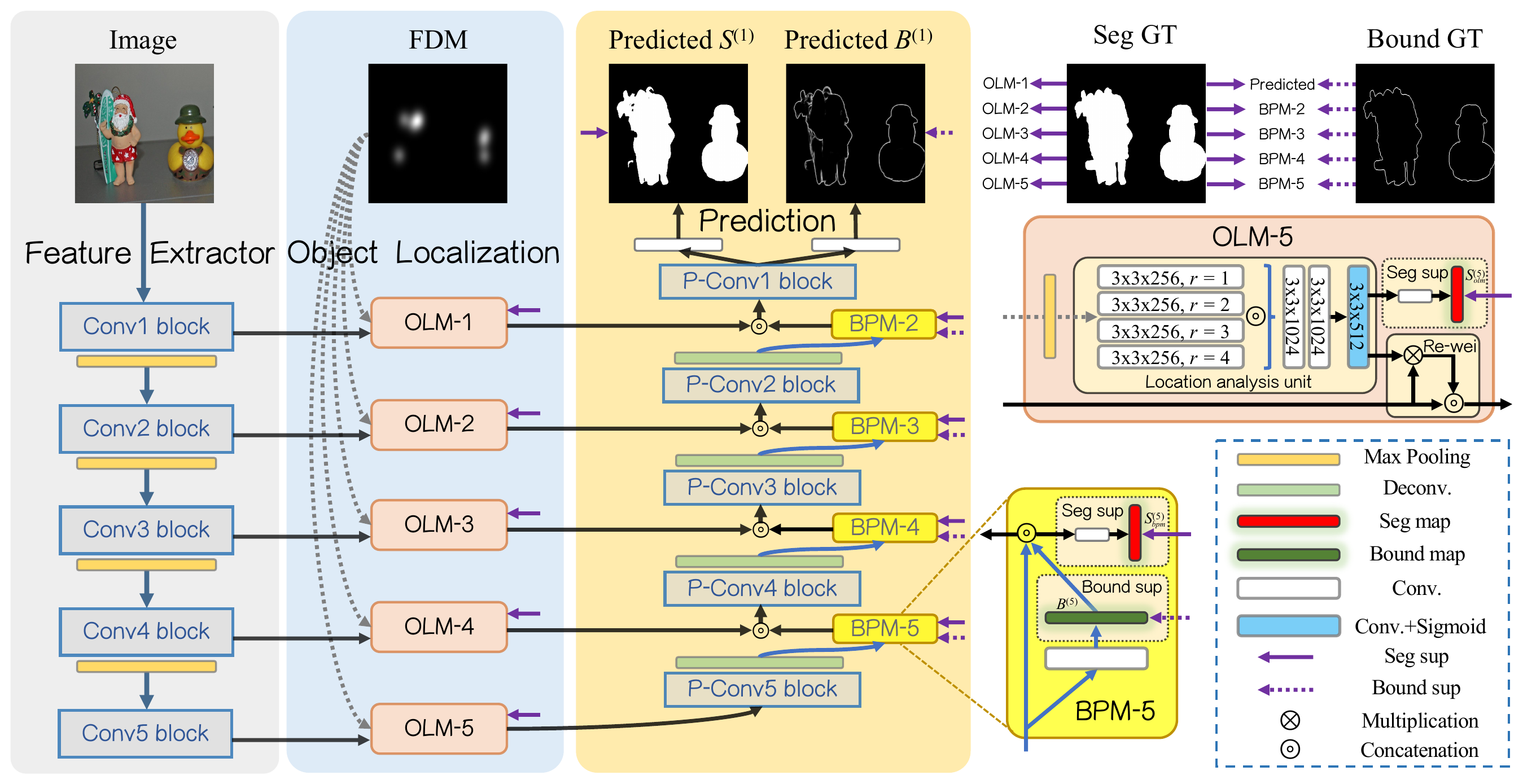}
\caption{The overall architecture of the proposed OLBP network.
OLBP network is organized in the mixed bottom-up and top-down manner.
We employ the modified VGG-16 to extract five blocks of features from an input image.
Then in each OLM, FDM is analyzed by several dilated and normal convolutional
layers to determine the location of objects in the corresponding block features.
Based on the object localization in each feature block, the top-down prediction is established.
During the prediction process, the boundary information is introduced into BPMs to guard
the completeness of objects and to filter background of erroneous localization.
We also construct a multi-task prediction structure, which contains object
segmentation branch and boundary prediction branch, to exploit the complementarity between
regions and boundaries.
}
\label{Fig3_Network_Overview}
\end{figure*}

\subsection{Data Preprocessing}
\label{sec:FDM}
The fixation points in each fixation map are sparse.
With only a few pixels per fixation map, there is too little valuable information to supply.
The similar problem arises in the clicks-based interactive image segmentation.
Xu \etal\cite{DIOS2016} transformed the clicks into Euclidean distance maps.
Inspired by this, we employ the Gaussian blur to transform the sparse fixation map
(\ie FM) into the \textit{fixation density map} (\ie FDM):
%
\begin{equation}
\label{FDM}
\mathbf{FDM}={\mathrm{nor}_\mathrm{min-max}}(\mathbf{FM} \circledast G_{\sigma}(x,y; \sigma)),
\end{equation}
where ${\mathrm{nor}_\mathrm{min-max}}(\cdot)$ is the min-max normalization,
$\circledast$ denotes convolution operator,
and $G_{\sigma}(\cdot)$ is a Gaussian filter with parameter $\sigma$
which is the standard deviation.
$\sigma$ is set corresponding to 1$^{\circ}$ visual angle in the OSIE dataset~\cite{OSIE2014}.
It is 24 pixels of an $800\times600$ image by default.

The effect of Gaussian blur is similar to the receptive field of eye, that is,
the center of fixation is with a high resolution and the surrounding of fixation is with a low resolution.
Thus, after performing Gaussian blur and linear transformation on FM,
the dense FDM contains more prior information of objects.
In this paper, we adopt the dense FDM rather than the raw FM.
We present an image with the personal fixations of three subjects of the PFOS dataset in
Fig.~\ref{Fig2_Examples}.
The fixation maps of Subject A and Subject B are constrained fixation maps,
while the fixation map of Subject C is an unconstrained fixation map.

\subsection{Network Overview and Motivation}
\label{Network Overview}
The proposed OLBP network has three critical components: the feature extractor,
the object locator and the prediction network with boundary preservation.
The overall architecture of OLBP network is illustrated in Fig.~\ref{Fig3_Network_Overview}.

\noindent\textbf{Feature Extractor.}
In the OLBP network, we adopt the modified VGG-16~\cite{2014VGG16}, from which
the last three fully connected layers have been removed, as the feature extractor.
We denote its input image as ${\bf I}\!\in\!\mathbb{R}^{H\!\times\!W\!\times\!C}$,
and initialize its parameters by the image classification model~\cite{2014VGG16}.
The feature extractor has five convolutional blocks, as shown in Fig.~\ref{Fig3_Network_Overview}.
We operate on the feature map of the last convolutional layer in each block,
\ie \textit{conv1-2}, \textit{conv2-2}, \textit{conv3-3}, \textit{conv4-3} and \textit{conv5-3},
which are denoted as $\{{\bf F}^{(i)}_{r}$: ${\bf F}^{(i)}_{r}\!\in\!\mathbb{R}^{\textit{h}_i\!\times\!
\textit{w}_i\!\times\!\textit{c}_i}, i=1,2,...,5\}$.
Notably, the feature resolution at the \textit{i-th} block, \ie $[\textit{h}_i,\textit{w}_i]$, is 
$[\frac{\textit{H}}{2^{i-1}},\frac{\textit{W}}{2^{i-1}}]$ and
$\textit{c}_{i\in\{1, 2, 3, 4, 5\}} = \{64, 128, 256, 512, 512\}$.
In reality, the input resolution $[\textit{H},\textit{W},\textit{C}]$ of ${\bf I}$ is set to
$288\times288\times3$.

\noindent\textbf{Object Localization Module.}
Although FDM is a probability map,
it is a critical interaction that reflects the intention of the user.
It is important to effectively explore the object location information of FDM.
However, when we construct a CNN-based model for the personal fixations-based object
segmentation task, it is natural to directly concatenate FDM and the input image for the network input.
Since there are three channels for image and only one channel for FDM,
the direct concatenation operation may drown out the critical interaction information of FDM.
Based on the above analysis, we propose the \textit{Object Localization Module} to process FDM.

The parallel convolution structure is effective to explore meaningful information in CNN features~\cite{TPAMI2018Deeplab},
especially with the dilated convolution~\cite{DilaConv2016}.
Thus, in OLM, we employ several parallel dilated convolutions with different dilation rates
to profoundly analyze the personal FDM to obtain object location information, which are a group of
response maps.
These response maps belong to ${[0,1]}^{\textit{h}_i\!\times\!\textit{w}_i\!\times\!\textit{c}_i}$,
which shows they have the same number of channels as the features of image at the \textit{i-th} block.
They are applied to re-weight features of image to highlight the gazed objects at channel-wise
and spatial-wise.
To enhance the location presentation of the response maps, we apply deep
supervision~\cite{2015DeepSup} in OLM.
As presented in Fig.~\ref{Fig3_Network_Overview}, the OLM is performed in a bottom-up manner,
and it is assembled after each block of feature extractor for strong object localization.
The detailed description of OLM is presented in Sec.~\ref{Object Localization Module}.
We show the ablation study of OLM in Sec.~\ref{Ablation Studies}, including a variant of
direct concatenation of image and FDM.

\noindent\textbf{Boundary Preservation Module and Prediction Network.}
Since some fixations fall in the background, there may be some noise on the
re-weighted feature of OLM. 
The ambiguity over the fixations causes great disturbance to the segmentation result.
Fortunately, there is \textit{a priori knowledge} that the background usually does not have a regular
boundary.
Thus, we introduce the boundary information into the prediction network, and propose the
\textit{Boundary Preservation Module} to filter the background of erroneous localization and preserve the
completeness of the gazed objects.
BPM is a momentous component to purify the segmentation result.
We also attach the pixel-level segmentation supervision and boundary supervision to BPM.
As shown in Fig.~\ref{Fig3_Network_Overview}, BPMs are equipped between convolutional
blocks in the prediction network from top to down.
To make full use of the boundary information, we also construct a multi-task structure in the prediction network.
We elaborate the formulation and ablation study of BPM in Sec.~\ref{Boundary Preservation Module}
and Sec.~\ref{Ablation Studies}, respectively.
%

\begin{table}[!t]
\centering
\caption{Detailed parameters of each OLM.
  We present the kernel size and channel number of each dilated/normal convolutions.
  Besides, we also present the dilation rates and the size of output feature.
  For instance, $(3\times3, 32)$ denotes that the kernel size is $3\times3$ and the channel number is 32.
  }
\label{tab:OLM_Details}
\renewcommand{\arraystretch}{1.3}
\renewcommand{\tabcolsep}{1.5mm}
\begin{tabular}{c|c|c|c|c}
\bottomrule
  \hline
  Aspects & \tabincell{c}{Dilation\\conv} & \tabincell{c}{Dilation\\rate} & 2$\times$Conv & Output size\\
  \hline
  \hline
    OLM-1 & $(3\times3, 32)$ & $1/3/5/7$ & $(7\times7, 128)$ & $[288\times288\times128]$  \\
    OLM-2 & $(3\times3, 64)$ & $1/3/5/7$ & $(5\times5, 256)$ & $[144\times144\times256]$  \\
    OLM-3 & $(3\times3, 128)$ & $1/3/5/7$ & $(5\times5, 512)$ & $[72\times72\times512]$  \\
    OLM-4 & $(3\times3, 256)$ & $1/2/3/4$ & $(3\times3, 1024)$ & $[36\times36\times1024]$  \\
    OLM-5 & $(3\times3, 256)$ & $1/2/3/4$ & $(3\times3, 1024)$ & $[18\times18\times1024]$  \\
\toprule
\end{tabular}
\end{table}

\subsection{Object Localization Module}
\label{Object Localization Module}
As the \textbf{OLM-5} shown in Fig.~\ref{Fig3_Network_Overview}, there are three main parts in the
\textit{Object Localization Module}: location analysis unit, feature re-weighting (\ie Re-wei) and
segmentation supervision (\ie Seg sup).
Its objective is to extract object location information of personal FDM
and to highlight objects in feature of image ${\bf F}^{(i)}_{r}$.
OLM is the most indispensable part of the whole OLBP network.

Concretely, in OLM-\textit{i}, the ${\bf FDM}\!\in\!\mathbb{R}^{H\!\times\!W\!\times\!1}$ is first
downsampled to fit the resolution of ${\bf F}^{(i)}_{r}$ and
to generate $\mathbf{F}^{(i)}_{fdm}\!\in\!\mathbb{R}^{\textit{h}_i\!\times\!\textit{w}_i\!\times\!1}$
which is formulated as:
\begin{equation}
   \begin{aligned}
    \mathbf{F}^{(i)}_{fdm} = \mathrm{MaxPool}(\mathbf{FDM}; {W}^{(i)}_{ks}),
    \label{eq:Pool}
    \end{aligned}
\end{equation}
where $\mathrm{MaxPool}(\cdot)$ is the max pooling with parameters ${W}^{(i)}_{ks}$,
which are $2^{i-1}\times2^{i-1}$ kernel with $2^{i-1}$ stride.

Then, we design the location analysis unit,
which contains four parallel dilated convolutions~\cite{DilaConv2016} with different dilation rates,
to analyze $\mathbf{F}^{(i)}_{fdm}$,
and obtain the multi-interpretation feature $\mathbf{F}^{(i)}_{mi}$.
The process in this unit can be formulated as:
\begin{equation}
   \begin{aligned}
    \mathbf{F}^{(i)}_{mi} = \mathrm{concat}\big( & C_{d}(\mathbf{F}^{(i)}_{fdm};{W}^{(i_1)}_{d}),C_{d}(\mathbf{F}^{(i)}_{fdm};{W}^{(i_2)}_{d}),\\
     & C_{d}(\mathbf{F}^{(i)}_{fdm};{W}^{(i_3)}_{d}),C_{d}(\mathbf{F}^{(i)}_{fdm};{W}^{(i_4)}_{d}) \big),
    \label{eq:Dilated}
    \end{aligned}
\end{equation}
where $\mathrm{concat}(\cdot)$ is the cross-channel concatenation,
and $C_{d}(\cdot ;{W}^{(i_n)}_{d})$ is the dilated convolution with
parameters ${W}^{(i_n)}_{d}$ for $n\in\{1,2,3,4\}$.
Notably, ${W}^{(i_n)}_{d}$ are comprised of kernel size, channel number and dilation rate.
Considering the resolution difference of each $\mathbf{F}^{(i)}_{r}$, the dilation rates of each unit
are different and the details are presented in Tab.~\ref{tab:OLM_Details}.
In this unit, the dilated convolutions large the receptive field without increasing the computation.
They are performed in a parallel manner, which makes $\mathbf{F}^{(i)}_{mi}$ effectively
capture the local and global location information of the gazed objects.

The multi-scale features in $\mathbf{F}^{(i)}_{mi}$ are complementary to each other.
They are blended to produce the location response maps
$\mathbf{r}^{(i)}_{loc}\!\in\!{[0,1]}^{\textit{h}_i\!\times\!\textit{w}_i\!\times\!\textit{c}_i}$ via:
\begin{equation}
   \begin{aligned}
           \mathbf{F}^{(i)}_{int} = {2C}(\mathbf{F}^{(i)}_{mi}; {W}^{(i)}_{2c}),
    \label{eq:Fusion1}
    \end{aligned}
\end{equation}
\begin{equation}
   \begin{aligned}
            \mathbf{r}^{(i)}_{loc} = \psi (C(\mathbf{F}^{(i)}_{int}; {W}^{(i)}_{c})),
    \label{eq:Fusion2}
    \end{aligned}
\end{equation}
where $\mathbf{F}^{(i)}_{int}$ is the interim feature,
$2C(\ast;{W}^{(i)}_{2c})$ are two convolutional layers with the same parameters ${W}^{(i)}_{2c}$,
$\psi(\cdot)$ is the sigmoid function,
and $C(\ast;{W}^{(i)}_{c})$ is the convolutional layer with parameters ${W}^{(i)}_{c}$ which are
$3\times3$ kernel with ${c}_{i}$ channels.
${W}^{(i)}_{2c}$ contain kernel size and channel number, which are different in different OLMs.
Their details are shown in the column with ``2$\times$Conv" of Tab.~\ref{tab:OLM_Details}.
%

\begin{figure}
\centering
\footnotesize
  \begin{overpic}[width=1\columnwidth]{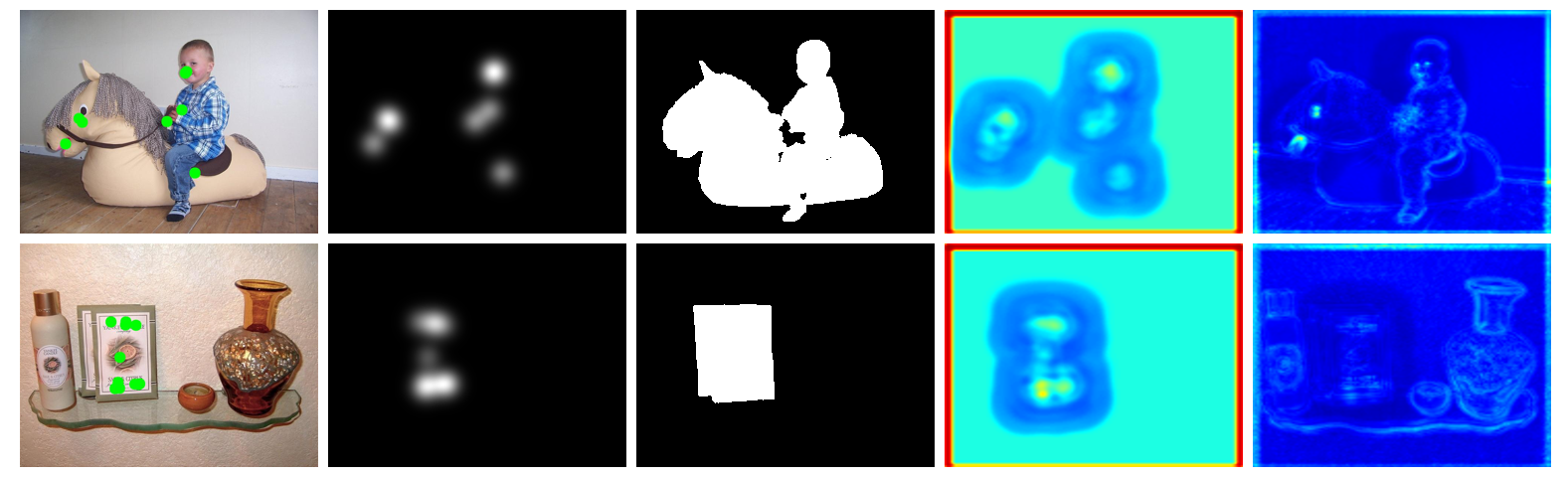}
    \put(5.9,-2.4){ Image }
    \put(25.9,-2.4){ FDM }
    \put(46.6,-2.4){ GT }
    \put(65.2,-3.1){ $\mathbf{r}^{(2)}_{loc}$ }
    \put(84.4,-3.1){ $\mathbf{F}^{(2)}_{loc}$}
  \end{overpic}
\caption{Feature visualization in OLM-2.
$\mathbf{r}^{(2)}_{loc}$ is the location response map,
and $\mathbf{F}^{(2)}_{loc}$ is the location-enhanced feature.
}
\label{Fig4_enhanced_Feature}
\end{figure}
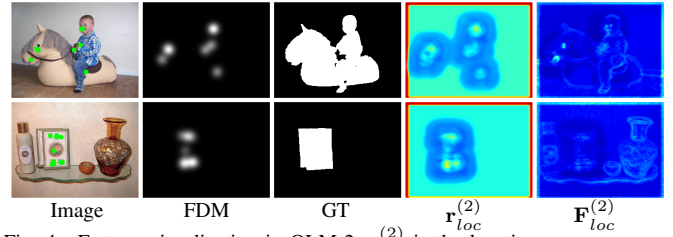

After completing the FDM interpretation in location analysis unit, we successfully
obtain $\mathbf{r}^{(i)}_{loc}$,
which are the protagonists of the feature re-weighting (\ie Re-wei) part.
We employ $\mathbf{r}^{(i)}_{loc}$ to re-weight $\mathbf{F}^{(i)}_{r}$ at channel-wise and
spatial-wise, and receive the location-enhanced feature
$\mathbf{F}^{(i)}_{loc}\!\in\!\mathbb{R}^{\textit{h}_i\!\times\!\textit{w}_i\!\times\!\textit{c}_i}$,
which is computed as:
\begin{equation}
   \begin{aligned}
      \mathbf{F}^{(i)}_{loc} = \mathbf{F}^{(i)}_{r} \otimes \mathbf{r}^{(i)}_{loc},
    \label{eq:Re-weight}
    \end{aligned}
\end{equation}
where $\otimes$ is element-wise multiplication.
Besides, in Re-wei, to balance the information of image and location, we concatenate
$\mathbf{F}^{(i)}_{r}$ to $\mathbf{F}^{(i)}_{loc}$ and obtain the output feature
$\mathbf{F}^{(i)}_{olm}$ of OLM.
The size of $\mathbf{F}^{(i)}_{olm}$ is shown in Tab.~\ref{tab:OLM_Details}.
Notably, at the training phase, we apply the pixel-level segmentation supervision
(\ie Seg sup) to each OLM.

In Fig.~\ref{Fig4_enhanced_Feature}, we visualize feature in
OLM-2 to verify the effectiveness of the location enhancement.
Concretely, in OLM-2, the \textit{conv2-2} is re-weighted by the location response map.
As shown in Fig.~\ref{Fig4_enhanced_Feature}, the location response map $\mathbf{r}^{(2)}_{loc}$
contains rich location information of the gazed objects.
After using Eq.~\ref{eq:Re-weight} to perform the location enhancement operation on \textit{conv2-2},
we observe that the gazed objects are highlighted in $\mathbf{F}^{(2)}_{loc}$ (with darker color).
In summary, the location-enhanced feature $\mathbf{F}^{(i)}_{loc}$ of OLM has strong location expression
ability and contributes to the subsequent segmentation prediction network.

\subsection{Boundary Preservation Module}
\label{Boundary Preservation Module}
The \textit{Boundary Preservation Module} is built to restrain the falsely highlighted part of the
re-weighted feature of OLM and to preserve the completeness of the gazed objects for the
segmentation prediction.
As the \textbf{BPM-5} shown in Fig.~\ref{Fig3_Network_Overview}, the structure of BPM is
succinct, but it is a key bridge to connect convolutional blocks of the prediction network.

Let $\{{\bf F}^{(i)}_{p}$: ${\bf F}^{(i)}_{p}\!\in\!\mathbb{R}^{\textit{h}_{i-1}\!\times\!
\textit{w}_{i-1}\!\times\!\textit{c}_{i-1}}, i=2,3,4,5\}$
denote the output feature of each deconvolutional layer in the prediction network.
In BMP, $\mathbf{F}^{(i)}_{p}$ is processed by a convolutional layer to generate the
boundary mask $\mathbf{B}^{(i)}$, which is defined as:
\begin{equation}
   \begin{aligned}
      \mathbf{B}^{(i)} = C(\mathbf{F}^{(i)}_{p}, {W}^{(i)}_{c}).
    \label{eq:Bound}
    \end{aligned}
\end{equation}

To increase the accuracy of $\mathbf{B}^{(i)}\!_{i\in\{2,3,4,5\}}$, we introduce the pixel-level
boundary supervision (\ie ``Bound sup" on \textbf{BPM-5} in Fig.~\ref{Fig3_Network_Overview})
in BPM.
Since that there are no pixel-level boundary annotations in the PFOS dataset,
we employ the morphological operation on binary segmentation GT $\mathbf{G}_s$
to produce the boundary GT $\mathbf{G}_b$, as follow:
\begin{equation}
   \begin{aligned}
     \mathbf{G}_{b} = \mathrm{Dilate}(\mathbf{G}_{s};\theta)-\mathbf{G}_{s},
    \label{eq:Dilate}
    \end{aligned}
\end{equation}
where $\mathrm{Dilate}(\ast;\theta)$ is the morphological dilation operation with dilation coefficient $\theta$
which is 2 pixels.

Then, $\mathbf{B}^{(i)}$ is concatenated to $\mathbf{F}^{(i)}_{p}$ to generate the output
feature $\mathbf{F}^{(i)}_{bpm}$ of BPM.
We also put the pixel-level segmentation supervision behind $\mathbf{F}^{(i)}_{bpm}$,
such as ``Seg sup" on \textbf{BPM-5} in Fig.~\ref{Fig3_Network_Overview}.
The segmentation supervision and the boundary supervision cooperate well with each other,
improving the feature representation of the gazed objects.
In this way, we novelly introduce boundary information into the BPM,
and $\mathbf{F}^{(i)}_{bpm}$ carries the feature de-noising and boundary preservation
capabilities into the prediction network.

\subsection{Implementation Details}
\label{Implementation Details}

\begin{table*}[t!]
  \centering
  \caption{Quantitative results including Jaccard index, S-measure, weighted F-measure,
E-measure and F-measure on the PFOS dataset (in percentage \%).
\textit{Semantic Segmentation} means semantic segmentation-based method.
\textit{Clicks} means clicks-based interactive image segmentation method.
\textit{Fixations} means fixations-based object segmentation method.
\textit{FDM-Guided Semantic Segmentation} means embedding FDM into semantic segmentation method.
\textit{FDM-Guided Salient Object Detection} means embedding FDM into salient object detection method.
The best three results are shown in \textcolor{red}{red}, \textcolor{blue}{blue},
and \textcolor{green}{green}.
$\uparrow$ denotes larger is better.
The subscript of each method represents the publication year.
$^\dagger$ means CNNs-based method.
}
\label{table:QuantitativeResults}
   \small
  \renewcommand{\arraystretch}{1.2}
  \renewcommand{\tabcolsep}{3.5mm}
\begin{tabular}{cc||ccccc}
\midrule[1pt]    
 \multirow{2}{*}{Aspects}  & \multirow{2}{*}{Methods} 
 & \multicolumn{5}{c}{PFOS Dataset} \\
 \cmidrule(l){3-7} 
           &  & $\mathcal{J} \uparrow$ & $\mathcal{S}_{\lambda} \uparrow$
             & $w\mathcal{F}_{\beta} \uparrow$
             & $\mathcal{E}_{\xi} \uparrow$ & $\mathcal{F}_{\beta} \uparrow$\\
\midrule[1pt]
 \multirow{6}{*}{\tabincell{c}{\textit{Semantic} \\ \textit{Segmentation} }}
 & PSPNet$_{17}$$^\dagger$~\cite{17PSPNet} & 51.0 & 58.9 & 55.5 & 64.2 & 60.2 \\
& SegNet$_{17}$$^\dagger$~\cite{TPAMI2017SegNet} & 58.7 & 70.4 & 66.6 & 78.4 & 72.5 \\
& DeepLab$_{18}$$^\dagger$~\cite{TPAMI2018Deeplab} & 52.8 & 65.7 & 60.5 & 72.9 & 66.9 \\
& EncNet$_{18}$$^\dagger$~\cite{EncNet2018} & 55.5 & 62.2 & 60.5 & 69.0 & 65.3 \\
& DeepLabV3+$_{18}$$^\dagger$~\cite{DeeplabV3} & 45.6 & 61.4 & 53.1 & 67.8 & 59.3 \\
& HRNetV2$_{19}$$^\dagger$~\cite{19HRNet} & 46.1 & 50.7 & 49.0 & 53.8 & 53.2   \\
\hline
 \multirow{3}{*}{\textit{Clicks}}
& ISLD$_{18}$$^\dagger$~\cite{LZW2018ISLD} & 61.2 & 73.4& 71.2 & 82.5 & 77.9 \\
& FCTSFN$_{19}$$^\dagger$~\cite{TSFN2019} & 62.4 & 72.9 & 69.9 & 82.8 & 75.1 \\
& BRS$_{19}$$^\dagger$~\cite{BRS2019} & 62.1 & 73.0 & 69.1 & 82.3 & 74.6 \\
\hline
 \multirow{4}{*}{\textit{Fixations}}	
& AVS$_{12}$~\cite{TPAMI2012AVS} & 40.9 & 56.0 & 48.7 & 65.1 & 56.6 \\
& SOS$_{14}$~\cite{2014SOS} & 42.6 & 57.5 & 51.4 & 67.8 & 60.0  \\
& GBOS$_{17}$~\cite{SR2017GBOS} & 38.0 & 56.7 & 48.0 & 63.9 & 58.1 \\
& CFPS$_{19}$$^\dagger$~\cite{LGY2019CFPS} & 70.5
                   & 78.9 & 76.7
                   & 87.4 & 81.3 \\
\hline
 \multirow{2}{*}{ \tabincell{c}{\textit{FDM-Guided} \\ \textit{Semantic Segmentation} }}
& DeepLabV3+$_{18}$$^\dagger$~\cite{DeeplabV3} & \textcolor{green}{\textbf{71.0}} 
	& \textcolor{green}{\textbf{79.5}} & \textcolor{green}{\textbf{78.3}} 
	&\textcolor{green}{\textbf{87.6}} & \textcolor{green}{\textbf{83.2}} \\
& HRNetV2$_{19}$$^\dagger$~\cite{19HRNet} & 58.8 & 71.3 & 68.6 & 80.4 & 75.7 \\
\hline
 \multirow{2}{*}{ \tabincell{c}{\textit{FDM-Guided} \\ \textit{Salient Object Detection} }}	
& CPD$_{19}$$^\dagger$~\cite{19CPD} & 69.2 & 78.4 & 76.4 & 86.2 & 81.7\\
& GCPA$_{20}$$^\dagger$~\cite{20GCPA} & \textcolor{blue}{\textbf{72.3}} 
			& \textcolor{blue}{\textbf{80.3}} & \textcolor{blue}{\textbf{78.9}} 
			& \textcolor{blue}{\textbf{88.1}} & \textcolor{blue}{\textbf{83.6}} \\
\hline
\hline
\textit{Personal Fixations} & \textbf{OLBP (Ours)} & \textcolor{red}{\textbf{73.7}} 
					& \textcolor{red}{\textbf{81.1}} & \textcolor{red}{\textbf{80.0}}
					& \textcolor{red}{\textbf{88.7}} & \textcolor{red}{\textbf{84.3}} \\
\toprule[1pt]
\end{tabular}
  \vspace{1pt}
  \vspace{-5pt}
\end{table*}

\noindent\textbf{Prediction Network.}
The prediction network is constructed in the top-down manner to gradually restore resolution.
It consists of five convolutional blocks, four BPMs and four deconvolutional layers.
A dropout layer~\cite{2014dropout} is placed before each deconvolutional layer to prevent
the prediction network from overfitting.
In addition, we attach the boundary prediction branch to the prediction network to assist
the object segmentation branch.
We initialize parameters of the prediction network by xavier method~\cite{2010Xavier}.

\noindent\textbf{Overall Loss.}
As shown in Fig.~\ref{Fig3_Network_Overview}, there are totally 15 losses in the OLBP network,
including 10 segmentation losses and 5 boundary losses.
The overall loss $\mathbb{L}$ can be divided into three parts: losses of multi-task prediction,
losses on OLMs and losses on BPMs.
$\mathbb{L}$ is calculated as:
\begin{equation}
   \begin{aligned}
    \mathbb{L}  = 
    & [\mathcal{L}_{s} (\mathbf{S}^{(1)},\mathbf{G}_s)+
    \mathcal{L}_{s} (\mathbf{B}^{(1)},\mathbf{G}_b)]+
    \sum\limits_{i=1}^5 \mathcal{L}_{s} (\mathbf{S}^{(i)}_{olm},\mathbf{G}_s)\\
    & +\sum\limits_{i=2}^5 [\mathcal{L}_{s} (\mathbf{S}^{(i)}_{bpm},\mathbf{G}_s)+
    \mathcal{L}_{s} (\mathbf{B}^{(i)},\mathbf{G}_b)],
    \label{eq:TotalLoss}
    \end{aligned}
\end{equation}
where $\mathcal{L}_{s} (\cdot,\cdot)$ is the softmax loss,
$\mathbf{S}^{(1)}$ is the predicted segmentation map,
and $\mathbf{B}^{(1)}$ is the predicted boundary map.
$\mathbf{S}^{(i)}_{olm}$ and $\mathbf{S}^{(i)}_{bpm}$ present the side
output segmentation results in OLM and BPM, respectively.
$\mathbf{B}^{(i)}\!_{i\in\{2,3,4,5\}}$ is the boundary mask in BPM.
Notably, for each softmax loss, we resize the resolutions of $\mathbf{G}_s$ and $\mathbf{G}_b$
to fit the resolutions of corresponding $\mathbf{S}^{(i)}_{olm}$, $\mathbf{S}^{(i)}_{bpm}$
and $\mathbf{B}^{(i)}$.

\noindent\textbf{Network Training.}
The PFOS dataset is separated into training set and testing set.
The training set contains 600 images with 9,000 personal fixation maps, including 3,075 constrained
fixation maps and 5,925 unconstrained fixation maps.
The testing set consists of 100 images with 1,500 personal fixations, including 608 constrained
fixation maps and 892 unconstrained fixation maps.

The OLBP network is implemented on Caffe~\cite{2014Caffe} and experimented using a NVIDIA Titan X GPU.
The data of training set and testing set are resized to $288\times288$ for training and inference.
We adopt the standard stochastic gradient descent (SGD) method~\cite{2010SGD} to
optimize our OLBP network for 30,000 iterations.
The learning rate is set to $8\times10^{-8}$, and it will be divided by 10 after 14,000 iterations.
The dropout ratio, batch size, iteration size, momentum and weight decay
are set to 0.5, 1, 8, 0.9 and 0.0001, respectively.

\section{Experiments}
\label{sec:exp}
In this section, we present comprehensive experiments on the proposed PFOS dataset.
We introduce evaluation metrics in Sec.~\ref{Evaluation Metrics}.
In Sec.~\ref{Comparison with the State-of-the-arts}, we compare the proposed OLBP network
with state-of-the-art methods.
Then, we conduct ablation studies in Sec.~\ref{Ablation Studies} and show some personal segmentation results in Sec.~\ref{Personal Segmentation Results}.
Finally, we present some discussions on the connections between fixation-based object segmentation and salient object detection in Sec.~\ref{Discussions}.

\subsection{Evaluation Metrics}
\label{Evaluation Metrics}
We use five evaluation metrics, \ie Jaccard index ($\mathcal{J}$),
S-measure ($\mathcal{S}_{\lambda}$)~\cite{Fan2017Smeasure},
F-measure ($\mathcal{F}_{\beta}$),
weighted F-measure ($w\mathcal{F}_{\beta}$)~\cite{2014WeiFm}, and
E-measure ($\mathcal{E}_{\xi}$)~\cite{Fan2018Emeasure},
to evaluate the performance of different methods.

\noindent\textbf{Jaccard Index} {$\mathcal{J}$}. Jaccard index is also called intersection-over-union (IoU),
which can compare similarities and differences between two binary maps.
It is defined as:
\begin{equation} \label{JaccardIndex}
\mathcal{J} = \frac{|\mathbf{S} \cap \mathbf{G}_s|}{|\mathbf{S} \cup \mathbf{G}_s|},
\end{equation}
where $\mathbf{S}$ is the predicted segmentation map, and $\mathbf{G}_s$ is the binary
segmentation GT.

\noindent\textbf{S-measure} $\mathcal{S}_{\lambda}$. 
S-measure focuses on the structural similarity between the predicted segmentation map
and the binary segmentation GT.
It evaluates the structural similarity of region-aware ($S_r$)
and object-aware ($S_o$) simultaneously.
S-measure is defined as:
\begin{equation} 
	\begin{aligned}
	\mathcal{S}_{\lambda} = \lambda \ast S_o + (1-\lambda) \ast S_r ,
	\label{S-measure}
	\end{aligned}
\end{equation}
where $\lambda$ is set to 0.5 by default.

\noindent\textbf{F-measure} $\mathcal{F}_{\beta}$.
F-measure is a weighted harmonic mean of precision and recall, which considers precision and recall comprehensively. It is defined as:
\begin{equation}
	\begin{aligned}
	\mathcal{F}_{\beta} = \frac{(1+\beta^2)\times Precision \times Recall}{\beta^2 \times Precision + Recall},
	\label{F-measure}
	\end{aligned}
\end{equation}
where $\beta^2$ is set to 0.3 following previous studies~\cite{2015SODBenchmark,2019sodsurvey}.

\noindent\textbf{Weighted F-measure} $w\mathcal{F}_{\beta}$.
Weighted F-measure has the ability to evaluate the non-binary and binary map. It focuses on evaluating the weights errors of predicted pixels according to their location and their neighborhood, which is formulated as:
\begin{equation}
	\begin{aligned}
	w\mathcal{F}_{\beta} = \frac{(1+\beta^2)\times Precision^w \times Recall^w}{\beta^2 \times Precision^w + Recall^w},
	\label{WeiF-measure}
	\end{aligned}
\end{equation}
where $\beta^2$ is set to 1 following previous studies~\cite{20ICNet,20CMWNet}.

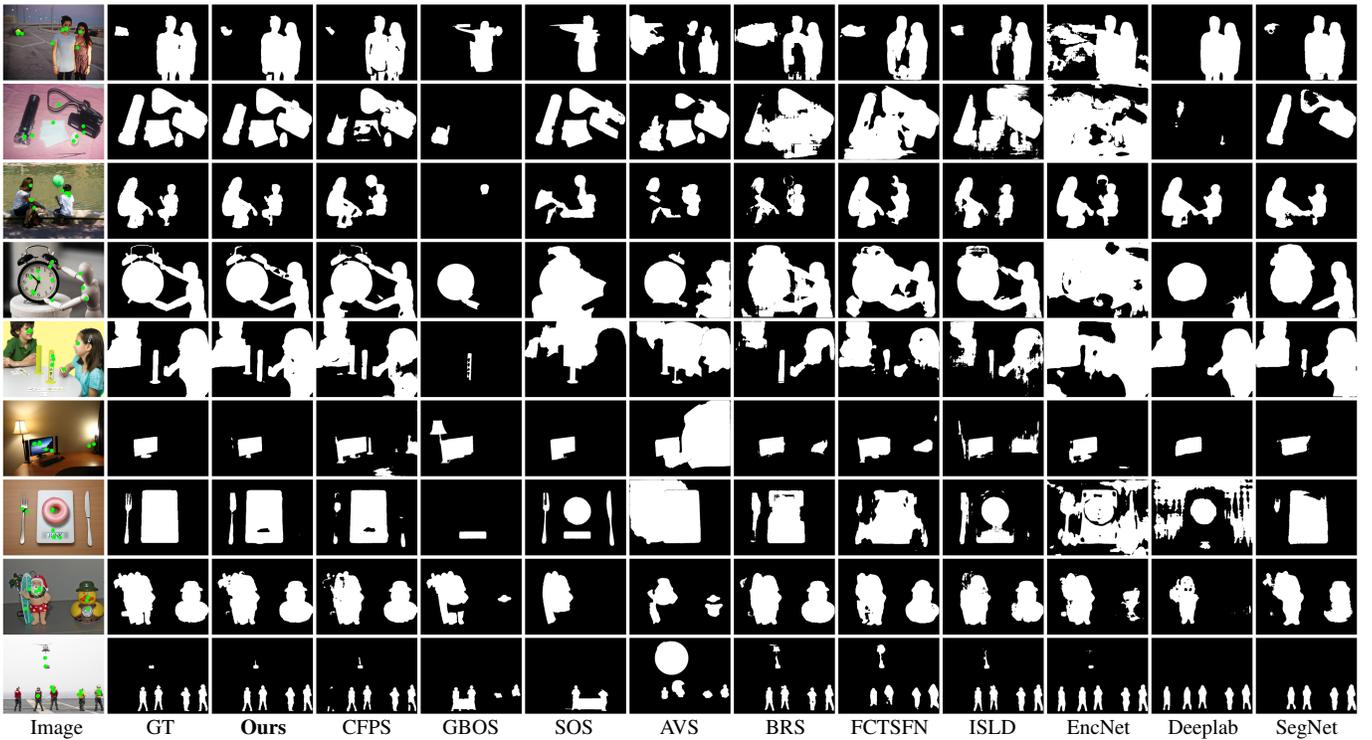
\begin{figure*}
\centering
\footnotesize
  \begin{overpic}[width=2.05\columnwidth]{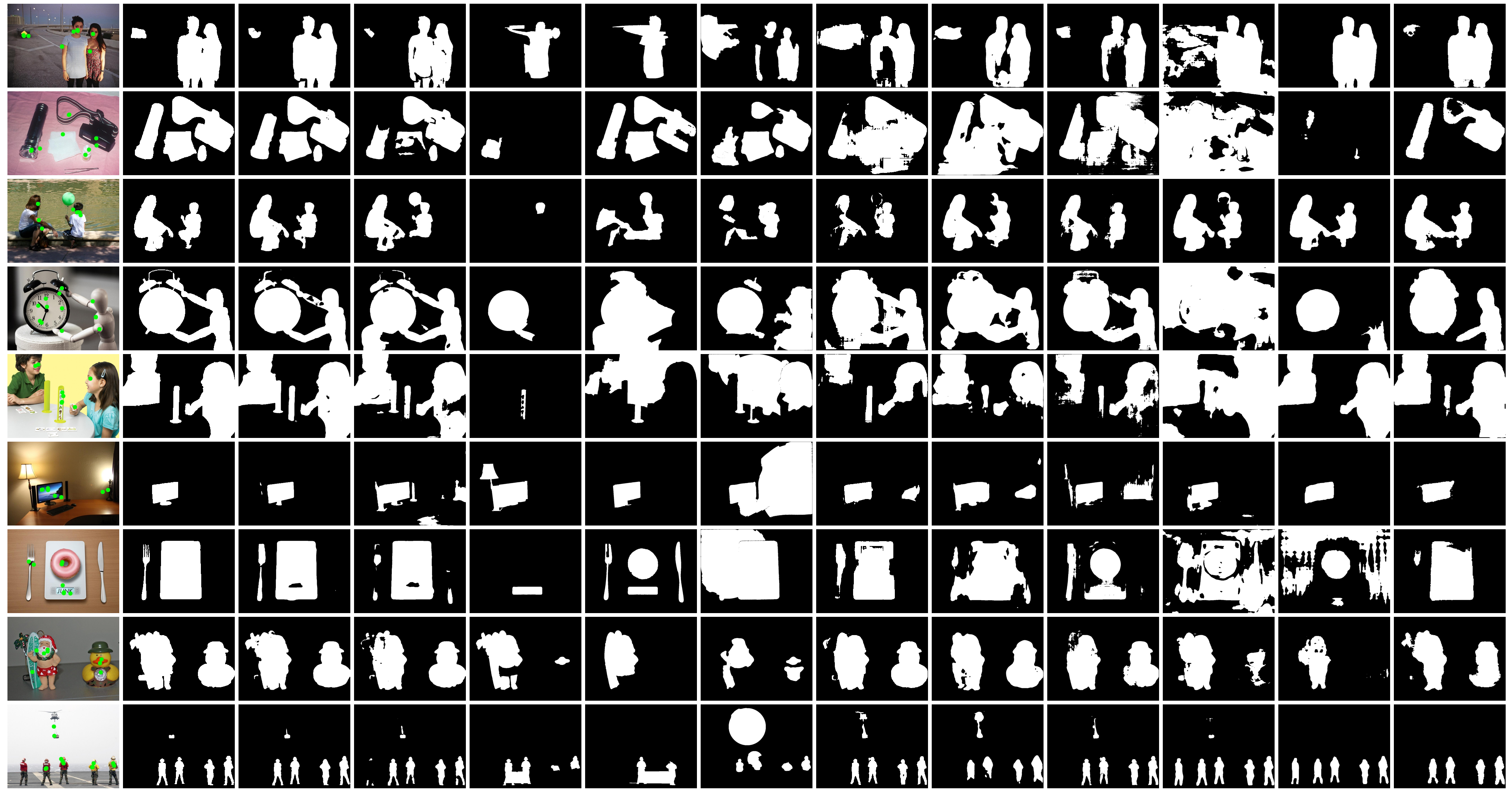}
    \put(1.9,-1.3){ Image}
    \put(10.4,-1.3){ GT }
    \put(17.3,-1.3){ \textbf{Ours} }
    \put(24.75,-1.3){ CFPS }
    \put(32.1,-1.3){ GBOS }
    \put(40.3,-1.3){ SOS }
    \put(48,-1.3){ AVS }
    \put(55.75,-1.3){ BRS }
    \put(62,-1.3){ FCTSFN }
    \put(70.65,-1.3){ ISLD }
    \put(77.8,-1.3){ EncNet }
    \put(85.2,-1.3){ Deeplab }
    \put(93.1,-1.3){ SegNet }
  \end{overpic}
\caption{Visualization comparison to some representative methods on the PFOS dataset.
Zoom-in for the best view.
}
\label{Fig_VisualRes}
\end{figure*}

\noindent\textbf{E-measure} $\mathcal{E}_{\xi}$.
E-measure is based on cognitive vision studies.
It evaluates the local errors (\ie pixel-level) and the global errors (\ie image-level) together.
We introduce it to provide a more comprehensive evaluation.
It could be computed as:
\begin{equation} 
	\begin{aligned}
	\mathcal{E}_{\xi} = \frac{1}{W\times H}\sum\limits_{x=1}^W \sum\limits_{y=1}^H f\Big(\frac{2\varphi_{\mathbf{G}_s} \circ \varphi_{\mathbf{s}}}{\varphi_{\mathbf{G}_s} \circ
	\varphi_{\mathbf{G}_s}+\varphi_{\mathbf{s}} \circ \varphi_{\mathbf{s}}}\Big),
	\label{E-measure}
	\end{aligned}
\end{equation}
where $\varphi_{\mathbf{G}_s}$ and $\varphi_{\mathbf{s}}$ are distance bias matrices
for binary segmentation GT and predicted segmentation map, respectively,
$\circ$ is the Hadamard product, and $f(\cdot)$ is the quadratic form.

\subsection{Comparison with the State-of-the-arts}
\label{Comparison with the State-of-the-arts}

\noindent\textbf{Comparison Methods.} 
We compare our OLBP network against three types of state-of-the-art methods, including 
\textit{semantic segmentation-based methods}, \textit{clicks-based interactive image segmentation methods} and 
\textit{fixations-based object segmentation methods}.
For a reasonable comparison of the first type of method, we follow~\cite{DIOS2016,LGY2019CFPS}, which
convert the segmentation problem into the selection problem.
Concretely, we first apply semantic segmentation methods, \ie 
PSPNet~\cite{17PSPNet},
SegNet~\cite{TPAMI2017SegNet},
DeepLab~\cite{TPAMI2018Deeplab},
EncNet~\cite{EncNet2018},
DeepLabV3+~\cite{DeeplabV3}, and 
HRNetV2~\cite{19HRNet}, to image, and then use the fixations to select the
gazed objects.
The second type of method includes ISLD~\cite{LZW2018ISLD}, FCTSFN~\cite{TSFN2019},
and BRS~\cite{BRS2019}.
The last type of method includes AVS~\cite{TPAMI2012AVS}, SOS~\cite{2014SOS},
GBOS~\cite{SR2017GBOS} and CFPS~\cite{LGY2019CFPS}.
For all the above compared methods, we use the implementations with recommend parameter settings
for a fair comparison.

In addition, we modify several semantic segmentation methods (\ie DeepLabV3+~\cite{DeeplabV3} and HRNetV2~\cite{19HRNet}) and recent salient object detection methods (\ie CPD~\cite{19CPD} and GCPA~\cite{20GCPA}) by embedding FDM in them to guide  object segmentation. Two types of comparison methods are thus generated, namely \textit{FDM-guided semantic segmentation} and \textit{FDM-guided salient object detection}, respectively.
Specifically, for DeepLabV3+, we embed FDM into features (\ie low-level features and features generated from the ASPP) to bridge the encoder and decoder; for HRNetV2, we embed FDM between the second stage and the third stage; for CPD, we embed FDM into two partial decoders; and, for GCPA, we embed FDM into four self refinement modules.
We retrain these modified methods with the same training dataset as our method, and their parameters are adjusted for better convergence.
Notably, we use the well-known OTSU method~\cite{OTSU} to binarize the generated probability map of our method and other CNNs-based methods.

\begin{table}[t!]
  \centering
  \caption{
Robustness evaluation of our method and several representative methods, such as the modified GCPA~\cite{20GCPA}, CFPS~\cite{LGY2019CFPS} and the modified CPD~\cite{19CPD}, on the test part of the PFOS dataset in terms of Jaccard Index.
The best result of each row is shown in \textbf{bold}.
Notably, ``+15\% noise'' means an additional 15\% increase in the number of unconstrained fixations of the total number of fixations in a fixation map.
We add the noise (\ie unconstrained fixations) at three levels, \ie 15\%, 30\%, and 45\%.
}
\label{table:RE}
   \small
  \renewcommand{\arraystretch}{1.25}
  \renewcommand{\tabcolsep}{2.8mm}
\begin{tabular}{c||ccccc}
\midrule[1pt]   
  & \textbf{OLBP} & GCPA$_{20}$ & CFPS$_{19}$ & CPD$_{19}$  \\     
                 
Dataset & \textbf{(Ours)}  & \cite{20GCPA} & \cite{LGY2019CFPS} & \cite{19CPD}  \\
\midrule[1pt]
PFOS  & \textbf{73.7} & 72.3 & 70.5 & 69.2 \\
\hline
+15\% noise  & \textbf{72.2} & 70.9 & 69.6 & 68.7  \\
+30\% noise  & \textbf{71.3} & 70.1 & 69.2 & 68.4  \\
+45\% noise  & \textbf{70.3} & 69.7 & 68.8 & 68.1  \\

\toprule[1pt]
\end{tabular}
\end{table}

\noindent\textbf{Quantitative Performance Evaluation.}
We evaluate our OLBP network and other 17 state-of-the-art methods on the PFOS dataset using
above five evaluation metrics.
The quantitative results are presented in Table~\ref{table:QuantitativeResults}.
Our OLBP network favorably outperforms all the compared methods in terms of different metrics.
Concretely, compared with the best method CFPS~\cite{LGY2019CFPS} in fixations-based
object segmentation methods, the performance
of our method is improved by $3.2\%$, $2.2\%$ and $3.0\%$ in $\mathcal{J}$, $\mathcal{S}_{\lambda}$
and $w\mathcal{F}_{\beta}$, respectively.
The performance of our method is $5.9\%$ better than FCTSFN~\cite{TSFN2019} in
$\mathcal{E}_{\xi}$, and is $6.4\%$ better than ISLD~\cite{LZW2018ISLD} in $\mathcal{F}_{\beta}$.
Note that the performance of our method is far better than that of three traditional methods
AVS~\cite{TPAMI2012AVS}, SOS~\cite{SR2017GBOS} and GBOS~\cite{SR2017GBOS}.
We attribute the performance superiority of the proposed OLBP network to the scheme of object localization and boundary preservation.

In addition, semantic segmentation-based methods get an average of $51.6\%$ in $\mathcal{J}$.
This may be due to the fact that semantic segmentation methods cannot accurately segment all
objects, resulting in the failure of the object selection process.
Clicks-based interactive image segmentation methods achieve an average of $61.9\%$ in $\mathcal{J}$,
while our OLBP network obtains $73.7\%$ in $\mathcal{J}$.
This demonstrates that our method is more robust than clicks-based interactive image segmentation
methods in adapting the ambiguity of fixations.
Fixations-based object segmentation methods contain three traditional methods and one
CNN-based method, obtaining an average of $48.0\%$ in $\mathcal{J}$. 

Specifically, we present the results of the FDM-guided semantic segmentation methods, including the modified DeepLabV3+ and HRNetV2, in Table~\ref{table:QuantitativeResults}.
The modified DeepLabV3+ achieves a promising performance, but does not exceed our OLBP network (\eg 71.0\% \textit{vs} 73.7\% in $\mathcal{J}$).
Although the FDM guidance brings some advantages to HRNetV2, but the modified HRNetV2 still does not perform well.
For the FDM-guided salient object detection, both modified CPD and GCPA perform well, though our OLBP still outperforms them (\eg 4.5\% and 1.4\% better than the modified CPD and GCPA in $\mathcal{J}$, respectively).
In summary, there is a large room for performance improvement on the proposed PFOS dataset, suggesting that the PFOS dataset is challenging to all compared methods including OLBP.

\begin{table}[!t]
\centering
\caption{Ablation analyses for the proposed OLBP network on the PFOS dataset (in percentage \%).
  As can be observed, each component in OLBP network plays an important role and contributes
  to the performance.
  The best result in each column is \textbf{bold}.
  Baseline: encoder-decoder network,
  OLM: object localization module,
  and BPM: boundary preservation module.
  }
\label{table:AblationStudyAll}
\renewcommand{\arraystretch}{1.4}
\renewcommand{\tabcolsep}{2.1mm}
\begin{tabular}{c|ccc||cccc}
\toprule
 & {Baseline} & {OLM} & {BPM} & $\mathcal{J} \uparrow$ & $\mathcal{S}_{\lambda} \uparrow$ & $w\mathcal{F}_{\beta} \uparrow$ \\
\midrule
1 &  \Checkmark$^{*}$ & &  & 67.2 & 75.9 & 72.2 \\ 

2 &  \Checkmark$^{*}$ & & \Checkmark & 68.0 & 76.4 & 72.5  \\
\hline
3 & \Checkmark & & & 70.7 & 78.3 & 75.0  \\ 
 
4 &  \Checkmark & \Checkmark &  & 73.0 & 80.7 & 79.5  \\

5 &  \Checkmark & & \Checkmark & 71.4 & 78.7 & 75.6  \\
\hline
6 &  \Checkmark & \Checkmark & \Checkmark & \bf{73.7} & \bf{81.1} & \bf{80.0} \\
\toprule
\multicolumn{7}{l}{\Checkmark$^{*}$ means the image and FDM are concatenated.}\\
\multicolumn{7}{l}{\Checkmark means the image and FDM are fed to network separately.}\\
\end{tabular}
\end{table}
\begin{figure}
\centering
\footnotesize
  \begin{overpic}[width=.99\columnwidth]{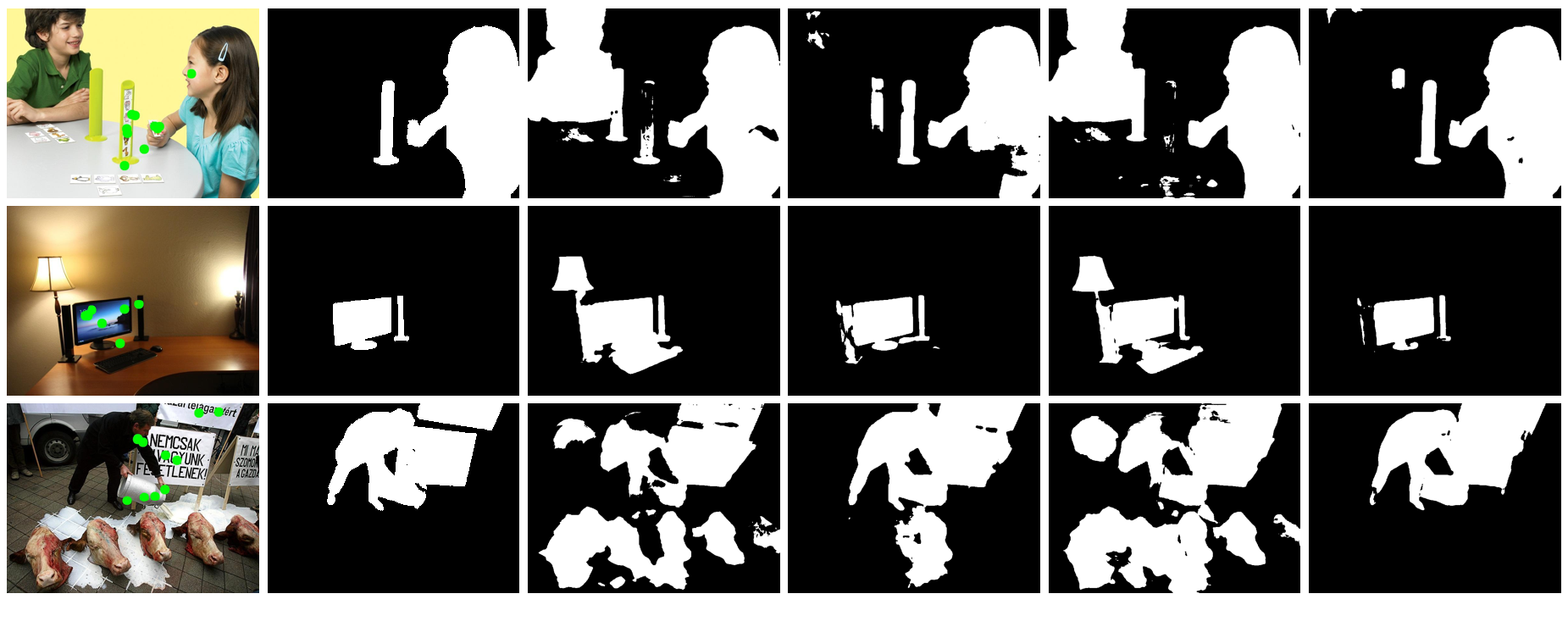}
    \put(4,-1){   Image  }
    \put(21.9,-1){ GT }
    \put(38.5,-1){   Ba$^{*}$  }
    \put(51.1,-1){ Ba+OLM }
    \put(66.65,-1){   Ba$^{*}$+BPM  }
    \put(87,-1){ Ours }
  \end{overpic}
\caption{Visual comparisons of different variants.
``Ba$^{*}$" is the baseline network, whose input is the concatenated image and FDM.  
}
\label{Fig_Ablation}
\end{figure}
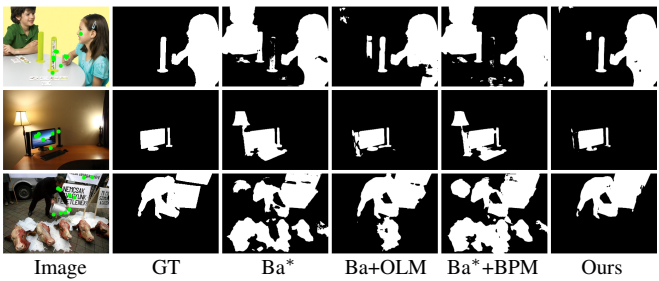

\noindent\textbf{Qualitative Performance Evaluation.}
In Fig.~\ref{Fig_VisualRes}, we show some representative visualization results of our OLBP network
and other methods.
Obviously, the visual segmentation maps of three traditional methods
GBOS~\cite{SR2017GBOS}, SOS~\cite{SR2017GBOS} and AVS~\cite{TPAMI2012AVS} are rough.
However, the CNN-based method CFPS~\cite{LGY2019CFPS}, which belongs to the same type as
GBOS, SOS and AVS, basically captures the gazed objects and brings in less background regions.
The gazed objects in the segmentation results of clicks-based interactive image segmentation methods
BRS~\cite{BRS2019}, FCTSFN~\cite{TSFN2019}, and ISLD~\cite{LZW2018ISLD} are
partially segmented and the details are relatively coarse.
As for the EncNet~\cite{EncNet2018}, DeepLab~\cite{TPAMI2018Deeplab} and SegNet~\cite{TPAMI2017SegNet}, the object segmentation maps of them depend on the semantic
segmentation results, which are great uncertainty.
This results in their object segmentation maps that are sometimes accurate and sometimes bad.

In contrast, our OLBP network is equipped with the scheme of object localization and boundary preservation,
which precisely analyzes the location information of fixations and completes the gazed objects.
The segmentation maps of ``Ours" in Fig.~\ref{Fig_VisualRes} are very localized in the gazed
objects with pretty fine details, even under the interference of some ambiguous fixations.

\noindent\textbf{Robustness Evaluation.}
We provide a robustness evaluation of our method and several representative methods, including the modified GCPA~\cite{20GCPA}, CFPS~\cite{LGY2019CFPS} and the modified CPD~\cite{19CPD}, on the test dataset of the PFOS dataset.
Concretely, we add the noise, \ie unconstrained fixations, to the fixation map by random sampling on the background regions at three levels, \ie different percentages (15\%, 30\%, 45\%) increase in the number of unconstrained fixations of the total number of fixations.
The performance of above methods after adding noise are presented in Table~\ref{table:RE}.
Our method consistently outperforms the compared methods under three challenging situations, showing excellent robustness.

\subsection{Ablation Studies}
\label{Ablation Studies}
We comprehensively evaluate the contribution of each vital component to performance in
our OLBP network.
Specifically, we assess
1) the overall contributions of OLM and BPM;
2) the effectiveness of the three parts in OLM; and
3) the usefulness of BPM and the top-down manner in prediction network.
The variants are retrained with the same hyper-parameters and training set as
aforementioned settings in Sec.~\ref{Implementation Details},
and the experiments are conducted on the PFOS dataset.

\begin{table}[!t]
\centering
\caption{Ablation results of the OLM on the PFOS dataset (in percentage \%).
  The best result in each column is \textbf{bold}.
  The corresponding structures of the listed variants are presented in Fig.~\ref{FigX_OLM}.
  }
\label{table:AblationStudy_OLM}
  \renewcommand{\arraystretch}{1.2}
  \renewcommand{\tabcolsep}{3.5mm}
\begin{tabular}{c||ccc}
\midrule[1pt]    
       OLM variants  & $\mathcal{J} \uparrow$ & $\mathcal{S}_{\lambda} \uparrow$ & $w\mathcal{F}_{\beta} \uparrow$ \\
\midrule[1pt]
\textit{w/o dilated convs} & 72.7 \tiny{-1.0}  & 80.7 \tiny{-0.4}  & 79.0 \tiny{-1.0} \\ 
\hline	  
\textit{w/o multiply}       & 72.6 \tiny{-1.1} & 80.5 \tiny{-0.6} & 79.2 \tiny{-0.8} \\
\textit{w/o concat}	   & 72.8 \tiny{-0.9} & 80.8 \tiny{-0.3} & 79.6 \tiny{-0.4} \\
\hline
\textit{w/o Seg sup}        & 72.9 \tiny{-0.8} & 80.9 \tiny{-0.2} & 79.4 \tiny{-0.6} \\
\hline
\hline
\textbf{Ours} & \textbf{73.7} & \textbf{81.1} & \textbf{80.0} \\
\toprule[1pt]
\end{tabular}
\end{table}
\begin{figure}
\centering
\footnotesize
  \begin{overpic}[width=.99\columnwidth]{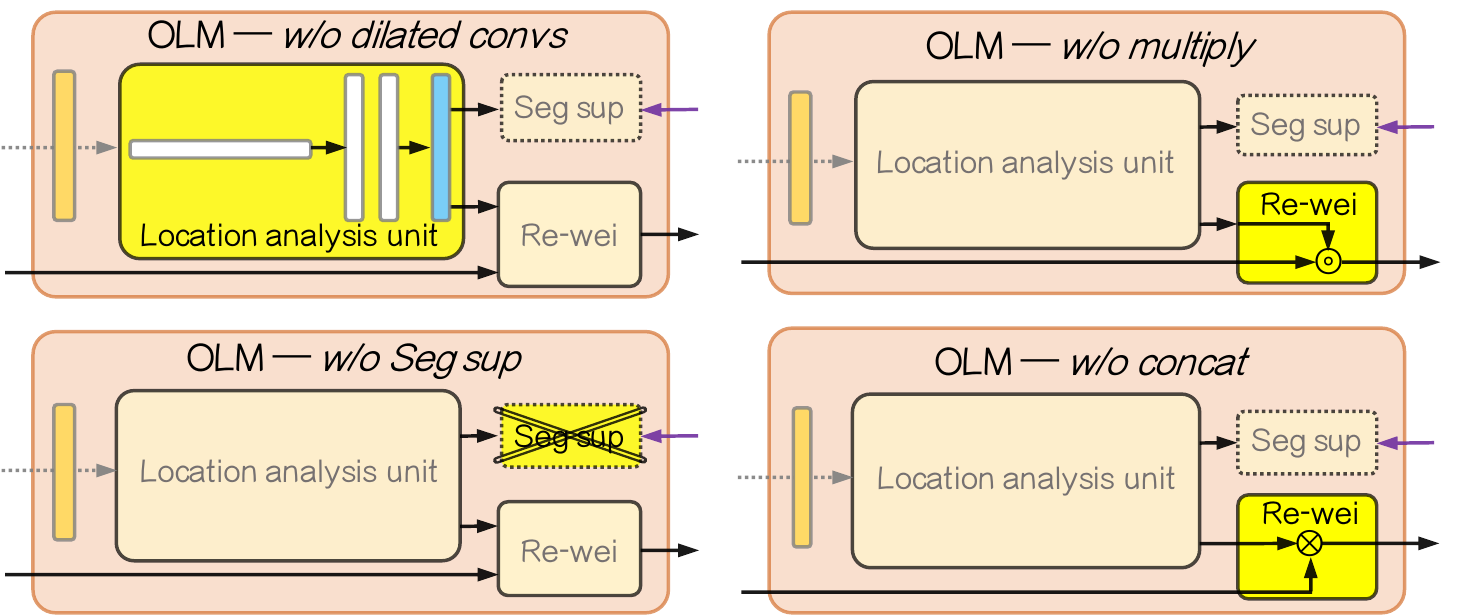}
  \end{overpic}
\caption{Structures of four OLM variants.
\textit{w/o dilated convs}: the four dilated convolutions are replaced by one convolutional layer;
\textit{w/o multiply}: without using response maps to re-weight image feature in Re-wei;
\textit{w/o concat}: without concatenating re-weighted feature and image feature in Re-wei;
\textit{w/o Seg sup}: without segmentation supervision.
}
\label{FigX_OLM}
\end{figure}

\begin{table}[!t]
\centering
\caption{The performance of side output segmentation maps of with/without BPM on PFOS
dataset (in percentage \%).
  The number in the lower right corner of the performance of w/o BPM is the difference between
  it and the performance of w/ BPM.
  The best result in each column is \textbf{bold}.
  }
\label{table:SideOutput}
  \renewcommand{\arraystretch}{1.2}
  \renewcommand{\tabcolsep}{1.6mm}
\begin{tabular}{c||ccc|ccc}
\midrule[1pt]    
 \multirow{2}{*}{Side outputs} 
 & \multicolumn{3}{c|}{\textit{w/ BPM} (\textbf{Ours})} & \multicolumn{3}{c}{\textit{w/o BPM}}\\
 \cmidrule(l){2-4} \cmidrule(l){5-7} 
             & $\mathcal{J} \uparrow$ & $\mathcal{S}_{\lambda} \uparrow$
             & $w\mathcal{F}_{\beta} \uparrow$
             & $\mathcal{J} \uparrow$ & $\mathcal{S}_{\lambda} \uparrow$
             & $w\mathcal{F}_{\beta} \uparrow$\\
\midrule[1pt]
$\mathbf{S}^{(5)}_{bpm}$ 	  & 62.0 & 71.6 & 67.8      & 60.2 \tiny{-1.8} & 70.4 \tiny{-1.2} & 66.0 \tiny{-1.8} \\
$\mathbf{S}^{(4)}_{bpm}$	  & 69.2 & 77.5 & 75.5      & 68.2 \tiny{-1.0} & 76.9 \tiny{-0.6} & 74.8 \tiny{-0.7} \\
$\mathbf{S}^{(3)}_{bpm}$ 	  & 72.6 & 80.2 & 78.9      & 71.9 \tiny{-0.7} & 79.8 \tiny{-0.4} & 78.3 \tiny{-0.6} \\
$\mathbf{S}^{(2)}_{bpm}$ 	  & 73.7 & 81.1 & 79.9      & 72.9 \tiny{-0.8} & 80.6 \tiny{-0.5} & 79.4 \tiny{-0.5} \\
\hline
\hline
$\mathbf{S}^{(1)}$ & \textbf{73.7} & \textbf{81.1} & \textbf{80.0}      & \textbf{73.0} \tiny{-0.7} & \textbf{80.7} \tiny{-0.4} & \textbf{79.5} \tiny{-0.5} \\
\toprule[1pt]
\end{tabular}
\end{table}

\textbf{1. Does the proposed OLM and BPM contribute to OLBP network?}
To evaluate the contribution of the proposed OLM and BPM to OLBP network,
we derive three variants: baseline network (denoted by ``Ba"/``Ba$^{*}$"), baseline network with only OLMs
(``Ba+OLM"), and baseline network with only BPMs (``Ba/Ba$^{*}$+BPM"). 
In particular, we provide two types of baseline network: the first one is an encoder-decoder network, whose input is the  concatenated image and FDM (denoted by ``Ba$^{*}$"); the second one is an encoder-decoder network with the down-sampled FDMs being concatenated to each skip-layer (denoted by ``Ba"), \ie the image and FDM are fed to network separately.
We report the quantitative results in Tab.~\ref{table:AblationStudyAll}.

We observe that the first baseline network ``Ba$^{*}$" (the 1$^{\mathrm{st}}$ line in Tab.~\ref{table:AblationStudyAll}) only obtains $67.2\%$ in $\mathcal{J}$, and the second baseline network ``Ba" (the 3$^{\mathrm{rd}}$ line in Tab.~\ref{table:AblationStudyAll}) obtains $70.7\%$ in $\mathcal{J}$.
This confirms that direct concatenation of the image and FDM results in the location information of FDM being submerged by image information; by contrast, concatenating FDM with image features at each scale benefits object location.
OLM significantly improves the performance of the baseline network (\eg $\mathcal{J}\!: 67.2\%/70.7\%\!\rightarrow\!73.0\%$ and $w\mathcal{F}_{\beta}\!: 72.2\%/75.0\%\!\rightarrow\!79.5\%$).
This shows that the contribution of OLM is remarkable, and OLM does capture the location information.
Comparing with OLM, the contribution of BPM to baseline networks is slightly inferior (\eg $\mathcal{J}\!: 67.2\%\!\rightarrow\!68.0\%$; $70.7\%\!\rightarrow\!71.4\%$), but BPM also shows its effectiveness to improve performance of ``Ba+OLM" (\eg $w\mathcal{F}_{\beta}\!: 79.5\%\!\rightarrow\!80.0\%$).
This demonstrates that BPM can further complete the objects and filter background of erroneous localization.
With the cooperation between OLM and BPM, the performance of the whole OLBP network is improved by $6.5\%/3.0\%$ in $\mathcal{J}$, $5.2\%/2.8\%$ in $\mathcal{S}_{\lambda}$ and $7.8\%/5.0\%$ in $w\mathcal{F}_{\beta}$ compared with the baseline network ``Ba$^{*}$"/``Ba".
This demonstrates that the scheme of bottom-up object localization and top-down boundary preservation is successfully embedded into the baseline network.

Additionally, the segmentation maps of variants based on the first baseline network ``Ba$^{*}$" are shown in Fig.~\ref{Fig_Ablation}.
We observe that ``Ba$^{*}$" almost segments all the objects in images.
With the assistance of OLM, ``Ba$^{*}$+OLM" determines the location of the gazed objects, and the
gazed objects on the segmentation maps of ``Ba+OLM" are much clearer.
Finally, with the help of BPM, the segmentation maps of ours (\ie OLBP network) are satisfactory.

\textbf{2. How effective are the three parts in OLM?}
As described in Sec.~\ref{Object Localization Module}, OLM consists of location analysis unit,
feature re-weighting (\ie Re-wei) and segmentation supervision (\ie Seg sup).
To validate the effectiveness of the three parts in OLM, we modify the structure of OLM and
provide four variants:
a) the four dilated convolutions are replaced by one convolutional layer in the location analysis
unit (\textit{w/o dilated convs});
b) without using response maps to re-weight image feature in Re-wei \textit{(w/o multiply});
c) without concatenating re-weighted feature and image feature in Re-wei (\textit{w/o concat}); and
d) without segmentation supervision (\textit{w/o Seg sup}).
The ablation results are reported in Tab.~\ref{table:AblationStudy_OLM}, and the detailed
structures of the above four OLM variants are presented in Fig.~\ref{FigX_OLM}.

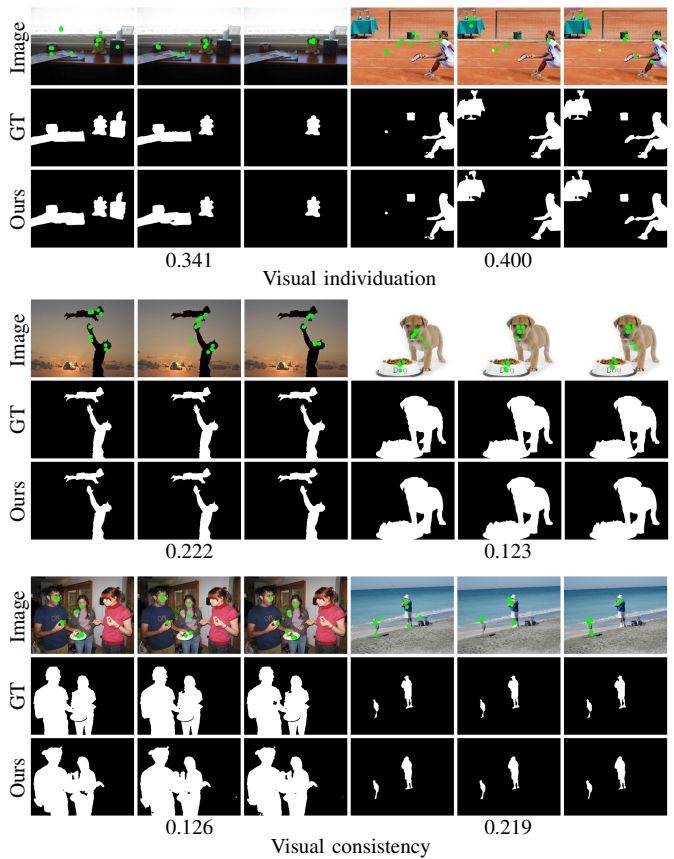
\begin{figure}
\centering
\footnotesize
  \begin{overpic}[width=.99\columnwidth]{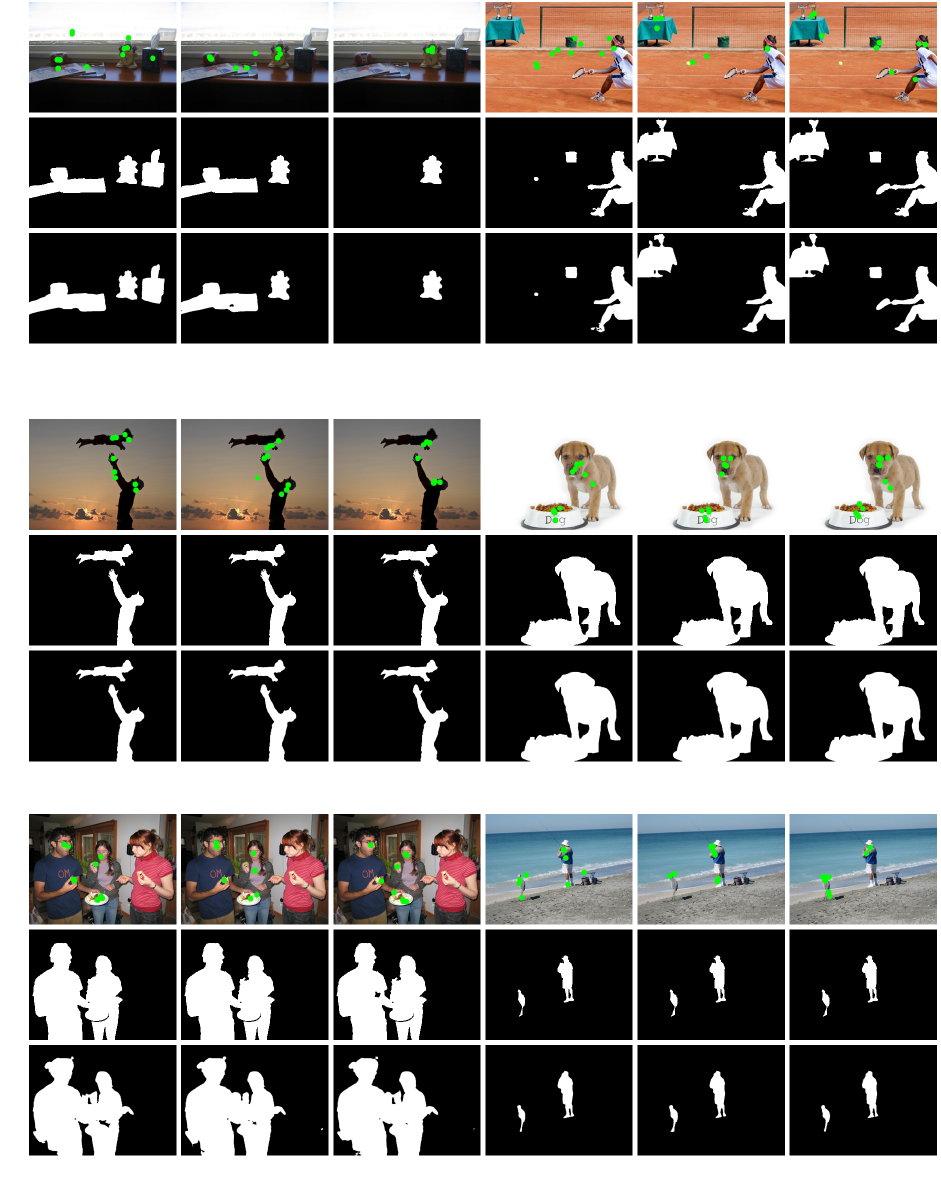}
    \put(29.5,66.3){  Visual individuation   }
    \put(17.8,68.5){  0.341   } \put(56.3,68.5){  0.400  }
    \put(17.8,33.3){  0.222  } \put(56.3,33.3){  0.123  }
    \put(30.5,-2.3){ Visual consistency }
    \put(17.8,-0.1){  0.126  } \put(56.3,-0.1){  0.219  }
    
    \put(-1,91.8){   \begin{sideways}{Image}\end{sideways}   }
    \put(-1,83.7){ \begin{sideways}{GT}\end{sideways}  }
    \put(-1,73.05){   \begin{sideways}{Ours}\end{sideways}   }
    
    \put(-1,56.45){   \begin{sideways}{Image}\end{sideways}   }
    \put(-1,48.35){ \begin{sideways}{GT}\end{sideways}  }
    \put(-1,37.7){   \begin{sideways}{Ours}\end{sideways}   }
    
    \put(-1,23){   \begin{sideways}{Image}\end{sideways}   }
    \put(-1,14.9){ \begin{sideways}{GT}\end{sideways}  }
    \put(-1,4.25){   \begin{sideways}{Ours}\end{sideways}   }
  \end{overpic}
\caption{Visual examples of personal segmentation results.
There are two basic properties of personal visual systems: visual individuation and
visual consistency.
The value of each image is the mean JS score.
}
\label{Fig_PVS}
\end{figure}

We discover that the performances of the four variants are worse than ours.
Concretely, the performance degradation of \textit{w/o dilated convs}
(\eg $\mathcal{J}\!: 73.7\%\!\rightarrow\!72.7\%$)
validates that the parallel dilated convolutions analyze FDM thoroughly and one
convolutional layer cannot mine sufficient location information from FDM.
The performance drop of \textit{w/o multiply}
(\eg $\mathcal{S}_\lambda\!: 81.1\%\!\rightarrow\!80.5\%$)
confirms that the location response maps
are more suitable to highlight objects on CNN feature of image than using them directly.
The reason behind this is that location response maps are a group of probability maps, without
rich object, texture and color information.
Besides, \textit{w/o concat} brings $0.9\%$ performance penalty in $\mathcal{J}$, which shows that
the information balance between image and location is important.
\textit{w/o Seg sup} carries $0.6\%$ performance drop in $w\mathcal{F}_\beta$.
This demonstrates that the segmentation supervision can enhance representation of the gazed objects.

\textbf{3. Is it useful to adopt BPM and the top-down manner in prediction network?}
To investigate the usefulness of the top-down manner in prediction network,
we report the performance of side output segmentation maps of BPM in
Tab.~\ref{table:SideOutput}.
Besides, we also report the side output performance of \textit{w/o BPM} in
Tab.~\ref{table:SideOutput} to evaluate the importance of BPM.

We observe that the quantitative results of side outputs ($\mathbf{S}^{(5)}_{bpm}$,
$\mathbf{S}^{(4)}_{bpm}$, $\mathbf{S}^{(3)}_{bpm}$, $\mathbf{S}^{(2)}_{bpm}$ and
$\mathbf{S}^{(1)}$) are incremental in terms of both
\textit{w/ BPM}
(\eg $w\mathcal{F}_\beta\!: 67.8\%\!\rightarrow\!75.5\%\!\rightarrow\!78.9\%\!\rightarrow\!79.9\%\!\rightarrow\!80.0\%$)
and \textit{w/o BPM}
(\eg $\mathcal{S}_\lambda\!: 70.4\%\!\rightarrow\!76.9\%\!\rightarrow\!79.8\%\!\rightarrow\!80.6\%\!\rightarrow\!80.7\%$).
This confirms that the top-down manner is useful for the prediction network.
The differences between the performance of \textit{w/o BPM} and \textit{w/ BPM} are also reported in
Tab.~\ref{table:SideOutput}.
We discover that all the differences are negative, which shows that BPM works well for each side
output of the top-down prediction network.

\subsection{Personal Segmentation Results}
\label{Personal Segmentation Results}
Due that the personal fixations are closely related to age and gender,
different users are interested in different objects when observing the same scene.
We define the visual difference of different personal visual systems as visual individuation.
Some examples of visual individuation are presented in the first part of Fig.~\ref{Fig_PVS}.
We can observe that there are multiple different types of objects and complex backgrounds
in these images.
The personal fixations of different users are located on different objects, which correspond to
the distinctive GTs.

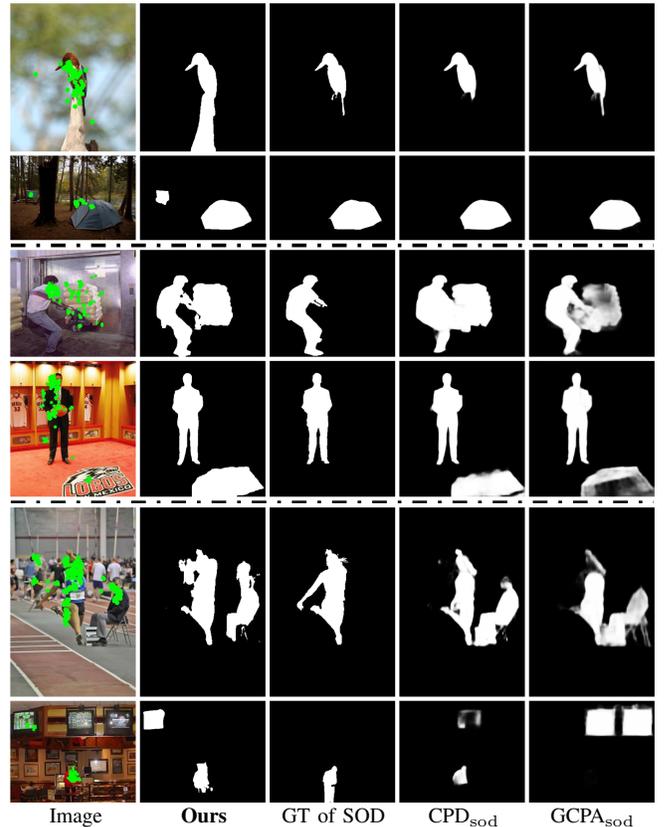
\begin{figure}
\centering
\footnotesize
  \begin{overpic}[width=.99\columnwidth]{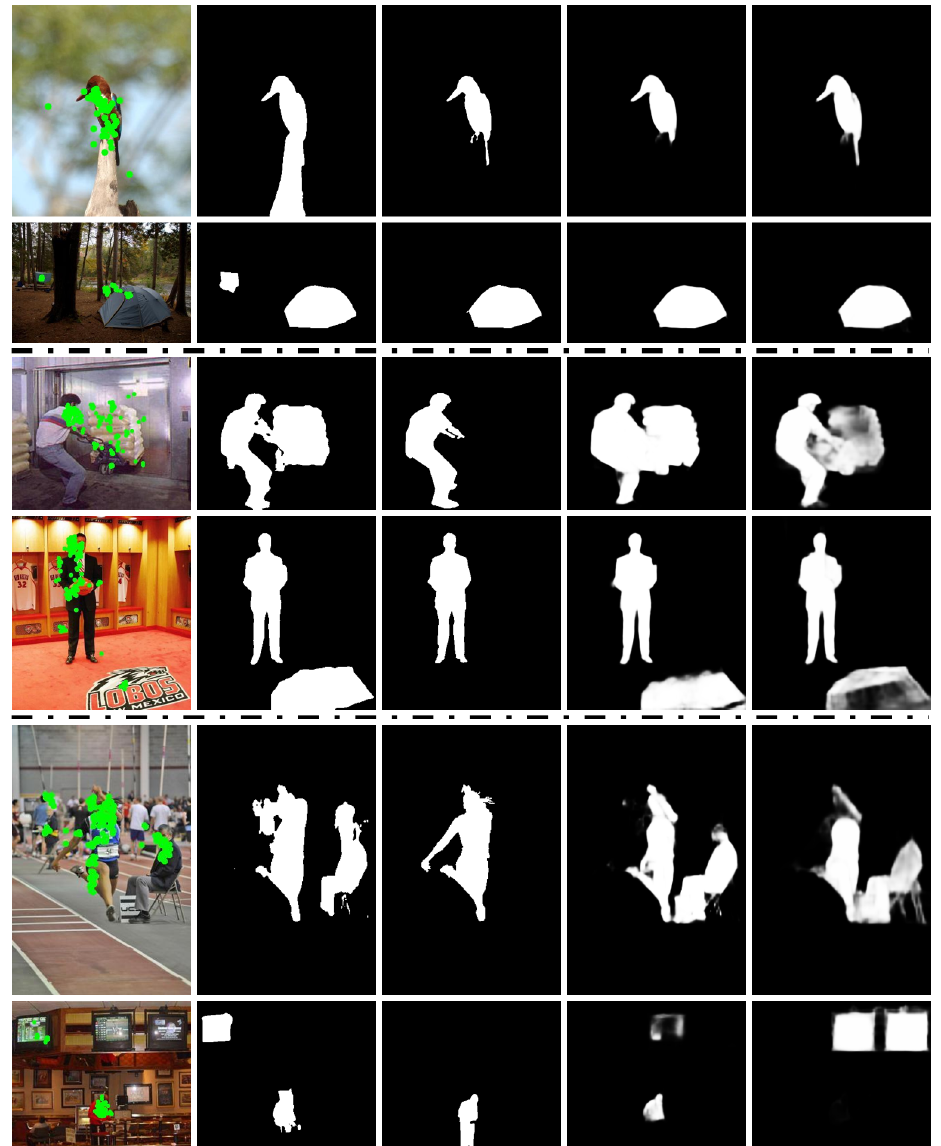}
    \put(4.9,-2.5){ Image}
    \put(21.3,-2.5){ \textbf{Ours}  }
    \put(34.8,-2.5){GT of SOD }
    \put(52.1,-2.5){ CPD$_\mathrm{sod}$ }
    \put(67.3,-2.5){ GCPA$_\mathrm{sod}$ }
  \end{overpic}
\caption{
Visual comparisons between our method, which is proposed for fixation-based object segmentation, and recent state-of-the-art salient object detection methods, including CPD~\cite{19CPD} and GCPA~\cite{20GCPA}, on the DUTS-OMRON~\cite{2013OMRON} and PASCAL-S~\cite{2014SOS} datasets.
``GT of SOD" means that the GT is for SOD task.
``CPD$_\mathrm{sod}$" means the original CPD method for SOD.
``GCPA$_\mathrm{sod}$" means the original GCPA method for SOD.
}
\label{Fig_9}
\end{figure}

In addition, we discover that personal visual systems are also
consistent in some scenes, which is denoted as visual consistency.
We show some examples of visual consistency in the second and third parts of Fig.~\ref{Fig_PVS}.
The images in the second part contain simple backgrounds and sparse objects, and the images in the third part contain more competitive situation, \ie complex background and partially selected objects.
In both parts, we observe that the locations of different personal fixations are similar, resulting in the identical GTs of different users.
Notably, in either case, our method show the ability to segment the gazed objects consistent with
the corresponding GT.

We also provide the quantitative analysis of visual individuation and visual consistency with Jensen-Shannon (JS) divergence.
JS divergence evaluates the similarity of two probability distributions $\mathbf{S}^1$ and $\mathbf{S}^2$, and it is based on Kullback-Leibler (KL) divergence. 
Its value belongs to [0,~1].
The closer its value is to zero, the smaller the difference between $\mathbf{S}^1$ and $\mathbf{S}^2$ is and the more similar they are.
It can be expressed as follows:
\begin{equation}
	\begin{aligned}
	\mathrm{JS}(\mathbf{S}^1,\mathbf{S}^2)=\frac{1}{2} \mathrm{KL}(\mathbf{S}^1,\frac{\mathbf{S}^1+\mathbf{S}^2}{2})+\frac{1}{2} \mathrm{KL}(\mathbf{S}^2,\frac{\mathbf{S}^1+\mathbf{S}^2}{2}),
	\label{JSD}
	\end{aligned}
\end{equation}
\begin{equation}
	\begin{aligned}
	\mathrm{KL}(\mathbf{P},\mathbf{Q})= \sum^{N}_{i=1} \mathbf{P}_\textit{i}\mathrm{log}\left(\epsilon +  \frac{\mathbf{P}_\textit{i}}{\epsilon+\mathbf{Q}_\textit{i}}\right),
	\label{KLD}
	\end{aligned}
\end{equation}
where KL($\cdot$) is Kullback-Leibler divergence, which is often used as an evaluation metric in fixation prediction~\cite{WWG2018DVA,LSTMFP2018TIP,LN2018FP,Che2020TIP}, \textit{i} indicates the \textit{i}$^{th}$ pixel in the probability distribution, $N$ is the total number of pixels, and $\epsilon$ is a regularization constant.

We introduce JS to measure the similarity of fixation points maps of each image in Fig.~\ref{Fig_PVS}.
First, we transform the fixation points map (green dots in each image) to FDM using Eq.~\ref{FDM};
then we compute the JS score of each two FDMs;
finally we report the mean JS score for each image in Fig.~\ref{Fig_PVS}.
It is obvious that the mean JS scores (\ie 0.222, 0.123, 0.126, and 0.219) of images which belong to visual consistency are relatively smaller than those (\ie 0.341 and 0.400) of images which belong to visual individuation.
And the mean JS scores of images which belong to visual consistency are close to zero, which indicates that the distributions of FDMs are very similar, \ie people may look at the same object(s).

\subsection{Discussions}
\label{Discussions}

Salient Object Detection (SOD) is widely explored in color images~\cite{19CPD,20GCPA,TIP15FKR,TMM19FKR,FKR2019NC}, RGB-D images~\cite{20Siamese,20JLDCF} and videos~\cite{ZXF2018TMM,Fan2019VideoSal,WWG19Video}, and it is closely related to our fixation-based object segmentation task.
In this section, we discuss the connections between fixation-based object segmentation and SOD.

SOD aims to highlight the most visually attractive object(s) in a scene, while fixation-based object segmentation aims to segment the gazed objects according to the fixation map, as defined in Sec.~\ref{Problem Statement}.
To illustrate the differences and connections between these two tasks, we conduct experiments on two SOD datasets, \ie DUTS-OMRON~\cite{2013OMRON} and PASCAL-S~\cite{2014SOS}, and show visual comparisons with two state-of-the-art SOD methods, \ie CPD~\cite{19CPD} and GCPA~\cite{20GCPA}, in Fig.~\ref{Fig_9}, which summarizes three situations.
First, in the $1^\mathrm{st}$ and $2^\mathrm{nd}$ rows, we present the differences of these two tasks: our method not only segments the salient objects, such as the bird and the big tent, but also segments the gazed wood stake and cloth that are not found in the GT of SOD and the results of CPD and GCPA.
Second, in the $3^\mathrm{rd}$ and $4^\mathrm{th}$ rows, we find that the results of CPD and GCPA are similar to ours, but different from the GT of SOD.
This shows that to some extent, the results of SOD methods CPD and GCPA are consistent with the fixation maps, even if the fixation maps are not exploited in these methods.
Third, in the $5^\mathrm{th}$ and $6^\mathrm{th}$ rows, we can clearly observe that our results are consistent with the fixation points in images, while the other three maps are different.
This shows that different SOD methods may cause confusion in some complicated scenes, resulting in inaccurate saliency maps.

Furthermore, we find that the salient objects always appear in the results of our method, while there is ambiguity among different SOD methods, which may highlight different salient objects.
So, to improve the accuracy of different SOD methods, we believe that the fixation-based object segmentation can be a pre-processing operation for SOD to determine the salient object proposals.

\section{Conclusion}
\label{sec:con}
In this paper, we propose a three-step approach to transform the available fixation prediction dataset
OSIE to the PFOS dataset for personal fixations-based object segmentation.
The PFOS dataset is meaningful to promote the development of fixations-based object segmentation.
Moreover, we present a novel OLBP network with the scheme of bottom-up object localization
and top-down boundary preservation to segment the gazed objects.
Our OLBP network is equipped with two essential components: the object localization module
and the boundary preservation module.
OLM is object locator, which is in charge of location analysis of fixations and object enhancement.
BPM emphasizes erroneous localization distillation and object completeness preservation.
Besides, we provide comprehensive experiments of our OLBP network and other three types
of methods on the PFOS dataset, which demonstrate the excellence of our OLBP network and
validate the challenges of the PFOS dataset.
In our future work, we plan to apply the proposed OLBP network to some eye-control devices,
facilitating the lives of patients with hand disability, ALS and polio.
In addition, we plan to recruit subjects to collect fixation points and corresponding ground truths on the PASCAL VOC~\cite{2010PASCALVOC} and MS COCO~\cite{MSCOCO} datasets for further exploring personal fixation-based object segmentation.




\ifCLASSOPTIONcaptionsoff
  \newpage
\fi

\bibliographystyle{IEEEtran}
\bibliography{ref}

%



%

\end{document}